\documentclass{article}
\usepackage[title]{appendix}
\usepackage {appendix}
\usepackage{PRIMEarxiv}
\usepackage[accsupp]{axessibility} 
\usepackage[utf8]{inputenc} 
\usepackage[T1]{fontenc}    
\usepackage{hyperref}       
\usepackage{cleveref}
\usepackage{url}            
\usepackage{booktabs}       
\usepackage{amsfonts}       
\usepackage{nicefrac}       
\usepackage{microtype}      
\usepackage{lipsum}
\usepackage{fancyhdr}       
\usepackage{graphicx}       
\usepackage{wrapfig}
\graphicspath{{media/}}     
\usepackage{subcaption}
\title{3MOS: Multi-sources, Multi-resolutions, and Multi-scenes dataset for Optical-SAR image matching
}

\author{
  Yibin Ye, Xichao Teng, Shuo Chen, Yijie Bian, Zhang Li\thanks{*Corresponding author} \\
  National University of Defense Technology \\
  \texttt{zhangli\_nudt@163.com*} \\
   \And
  Tao Tan\\
  Macao Polytechnic University \\
  \texttt{taotanjs@gmail.com} \\
}

\begin{document}
\maketitle

\begin{figure}[h]
  \centering
  \includegraphics[height=9cm]{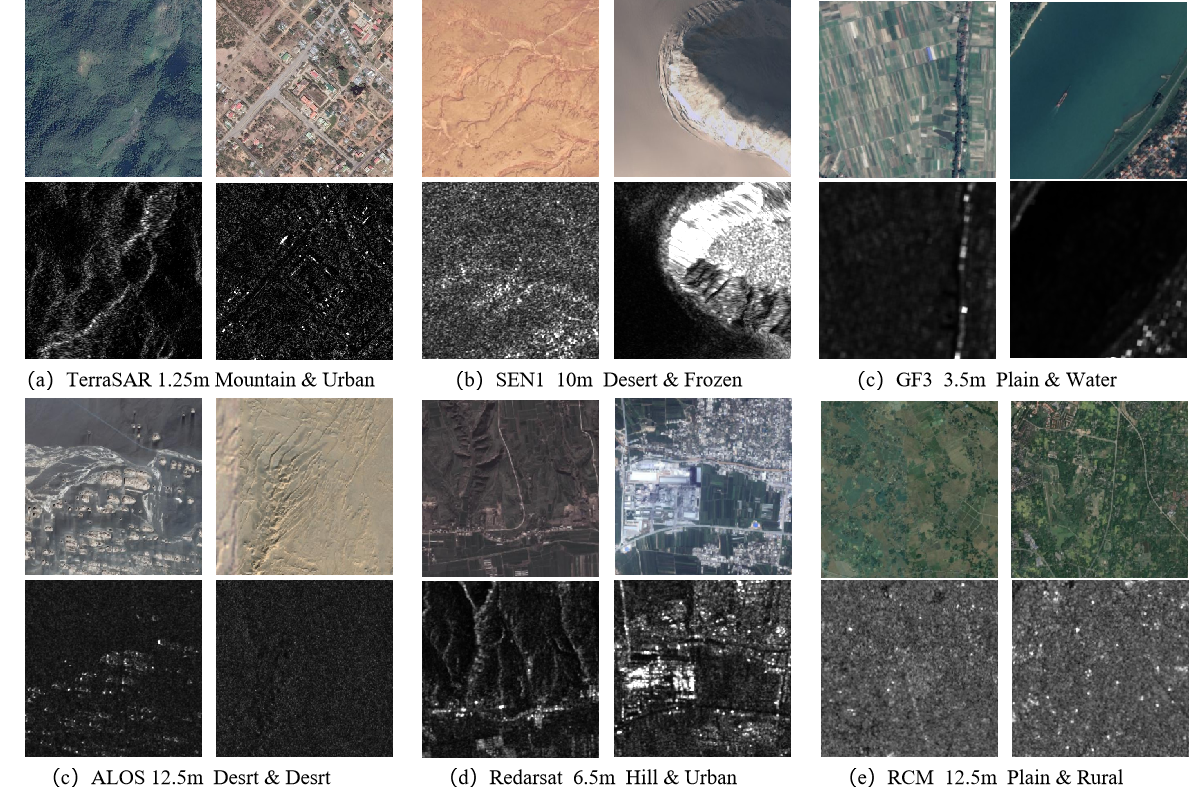}
  \caption{ Overview of 3MOS dataset. 3MOS contains 155K optical-SAR image pairs, including
 SAR data from 6 commercial satellites. The data is  registered and classified into 8 scenes including  urban, rural, plains, hills, mountains, water, desert, and frozen earth.
  }
  \label{fig:overview}
\end{figure}
\begin{abstract}
Optical-SAR image matching is a fundamental task for image fusion and visual navigation.  However, all large-scale open SAR dataset for methods development are collected from single platform, resulting in \textbf{limited} satellite types and spatial resolutions. Since images captured by different sensors vary significantly in both geometric and radiometric appearance, existing methods may fail to match corresponding regions containing the same content. Besides, most of existing datasets have not been categorized based on the characteristics of \textbf{different scenes}. To encourage the design of more general multi-modal image matching methods, we introduce a large-scale \textbf{M}ulti-sources, \textbf{M}ulti-resolutions, and \textbf{M}ulti-scenes dataset for \textbf{O}ptical-\textbf{S}AR image matching(\textbf{3MOS}). It consists of 155K optical-SAR image pairs, including SAR data from \textbf{six commercial satellites}, with resolutions ranging from \textbf{1.25m to 12.5m}. The data has been classified into \textbf{eight scenes} including urban, rural, plains, hills, mountains, water, desert, and frozen earth. Extensively experiments show that none of state-of-the-art  methods achieve consistently superior performance across different sources, resolutions and scenes. In addition, the distribution of data has a substantial impact on the matching capability of deep learning models, this proposes the domain adaptation challenge in optical-SAR image matching. Our data and code will be available at: \url{https://github.com/3M-OS/3MOS}.
\end{abstract}


\section{Introduction}
Image matching, as a fundamental task in computer vision, refers to identifying and then corresponding the same or similar structure from different images\cite{JiayiMa, XingyuJiang}. With the development of deep learning, the accuracy and robustness of image matching methods have significantly improved\cite{Dusmanu2019CVPR, detone18superpoint, Pautrat_2020_ECCV, he2023darkfeat, lightglue, zhang2023convmatch}. However, most of the deep learning matching research is based on ground-based unimodal image datasets like Hpatches\cite{Hpatches, lightglue}, MegaDepth\cite{MegaDepth, chen2022aspanformer}, etc. and there is still relatively few deep neural networks designed for multimodal remote sensing images matching(MRSIM), one important reason is the scarcity of high-quality MRSIM datasets\cite{BaiZhu}. MRSIM is increasingly crucial because it is a prerequisite and pivotal step in plentiful co-processing and integration applications in the field of remote sensing, such as image fusion\cite{Zhao_2023_CVPR}, visual navigation\cite{DLKFM}, and change detection\cite{ChangeDetection1}. To encourage deep learning based MRSIM research, we provide a \textbf{M}ulti-sources, \textbf{M}ulti-resolution, and \textbf{M}ulti-scene \textbf{O}ptical-\textbf{S}AR image matching dataset (\textbf{3MOS}).

Optical-SAR data fusion holds profound significance in the field of remote sensing due to the stark differences in geometric and radiometric appearance\cite{Fusion1,Fusion2,Fusion3,Fusion4}. Optical images, relying on angular measurements, provide information about the chemical properties of the observed environment.\cite{SEN12} Unfortunately, they are sensitive to cloud cover or fog, and ineffective for nighttime imaging. On the other hand, SAR images is based on range measurements and offer all-weather and all-day imaging capabilities. However, SAR images often contain much random noise, speckle noise, as well as specific phenomena resulting from imaging mechanisms such as top-down inversion and layover. Therefore, the structure and details of SAR images are more challenging to interpret compared to optical images\cite{Whu_SEN_city}. Although there are some available optical-SAR image matching datasets, they have limited satellite types and spatial resolutions, and they have not been categorized based on the characteristics of different scenes in optical and SAR images(\cref{tab:datasets}). In contrast, the 3MOS dataset integrates various satellites and various spatial resolutions. The data are also categorized into eight scenes based on ground features like mountains, plains, cities, water, deserts, etc., which can effectively enhance the diversity of training data and the specificity of test data. 

We expect 3MOS dataset to play a role in applications like multimodal image fusion and visual navigation:

—In the field of multi-sources satellite image fusion\cite{MultisourceFusion1, MultisourceFusion2,MultisourceFusion3}, a single satellite is difficult to achieve high-frequency observations of key areas. Meanwhile, different modal sensors have different imaging characteristics and advantages. At this point, fusing multimodal data can compensate for the insufficient observation capability of a single satellite. High-precision image matching is the foundation of this application. Due to significant differences in imaging modes and sensor parameters among different satellites, it is necessary to use data from different satellites for training and testing in order to improve the robustness and domain adaptation ability of the matching model.

— In the field of aircraft navigation, Matching real-time SAR images captured by the aircraft with pre-loaded optical reference images containing geo-referenced information can enable visual navigation\cite{VisualNavigation1}. On one hand, ground scenes vary greatly, with different features exhibited by different areas. High-precision visual navigation applications require matching algorithms to produce reliable results in different scenes. On the other hand, the spatial resolution of ground imaging by the aircraft varies at different altitudes, and the types of SAR sensors carried and the sources of optical reference images used may also differ significantly. Therefore, it is necessary to conduct simulation experiments using data from multiple scenes, resolutions, and sources.

To demonstrate the significance of constructing datasets using data from different satellites, resolutions, and scenes, we conducted experiments using traditional multimodal image matching methods and deep learning methods on the 3MOS dataset. We found that existing methods exhibit lower matching accuracy in challenging scenes such as deserts and mountainous areas compared to scenes with clear features like cities. Additionally, models trained on the same satellite data may experience a decrease in matching capability when applied to data from different satellites and resolutions. This suggests that improvements are needed in current multimodal image matching methods. In summary, our main contributions are as follows:

—We construct a multi-sources, multi-resolution, and multi-scene optical-SAR image matching dataset called 3MOS, which is challenging and meaningful for MRSIM problem.

—We use state-of-the-art methods to do template matching experiments in 3MOS dataset, and we find none of these methods achieve consistently superior performance across different sources, resolutions and scenes. 

—We introduce a multi-scale feature network called MFN as baseline and the cross-training and testing experiments have demonstrated the differences in data distributions. This proposes the domain adaptation problem in optical-SAR image matching.

\section{Related work}
\label{sec:Relate}
\subsection{Optical-SAR image matching dataset}
 Table \ref{tab:datasets} reports recent datasets for optical-SAR image matching. Although some studies have utilized other datasets for algorithm testing, these datasets have a small quantity of images. Each individual dataset only contains a few to several tens of images, hence they are not included in the table. From the table, it can be observed that the existing open-source datasets mainly consist of SAR images from a single sensor and optical images mostly from Google Earth or SEN2.  There are significant variations among SAR images acquired by different types of SAR sensors\cite{SOpatch}, yet there is no dataset encompassing SAR data from more than three satellites. It is worth noting that, despite SOPatch containing 650K image pairs utilizing SAR data from both Sentinel-1 and GF-3, the data in this dataset was reprocessed from WHU-SEN-City, SEN1-2, and os-dataset, and the image size is limited to 64x64. Additionally, the spatial resolution of existing datasets ranges from 0.5m to 10m, without a dataset simultaneously covering different resolutions. In terms of data scenes, some datasets only focus on urban areas (MSAW, SARptical). Although SEN1-2, WHU-SEN-City and SOPatch contain various scenes, they have not differentiate them. SEN12MS has provided land cover data for each image, but it does not distinguish the scenes for specific image pairs. Liao et al. divides the image pairs into six categories, but their dataset is not open-source. Therefore, there is currently a lack of matching datasets that divide scenes according to geographic characteristics.
\begin{table}[h]
  \caption{Comparing recent datasets for optical-SAR image matching. SR:spatial resolution. * not open-sourced.
  }
  \label{tab:datasets}
  \centering
  \scalebox{0.65}{
  \begin{tabular}{l c ccccp{1.5cm} p{1.8cm} p{4cm}}   
    \toprule
    name & Year & SR & SAR source & optical source &  scenes & image size & Number of \newline image pairs	& Image registration     \newline      method
\\
    \midrule
    SARptical\cite{SARptical} & 2017 & 1m & TerraSAR-X & aerial camera & urban & 112x112 & 10K & 3D reconstruction + \newline point cloud matching\\
QXS-SAROPT\cite{QXS-SAROPT} & 2021 & 1m & GF3 & Google Earth & urban & 256x256 & 20K & Manually registration \\
os-dataset\cite{osdataset} & 2020 & 1m & GF3 & Google Earth & urban and plains & 256x256 & 10.6K & PC-based method\cite{osdataset} \\
SEN1-2\cite{SEN12} & 2018 & 10m & SEN1 & SEN2 & multiple but mixed & 256x256 & 282K & unknown  \\
SEN12MS\cite{SEN12MS} & 2019 & 10m & SEN1 & SEN2 & multiple but mixed & 256x256 & 180.6K & ortho-rectified \\
WHU-SEN-City\cite{Whu_SEN_city} & 2019 & 10m & SEN1 & SEN2 & multiple but mixed & 256x256 & 18.5K & unknown \\
MSAW\cite{MSAW} & 2020 & 0.5m & aerial SAR & Maxar Worldview2 & urban & 900x900 & 6K & ortho-rectified \\ 
Liao et.al\cite{Matchosnet}* & 2022 & 5m & GF3 & SEN2 & multiple & 512x512 & 96K & unknown \\
SOPatch\cite{SOpatch} & 2023 & 1m/10m & GF3/SEN1 & SEN2 & multiple but mixed & 64x64 & 650K & RIFT and CFOG \\
\textbf{3MOS(ours)} & 2024 & multiple & multiple & Google Earth & multiple & 256x256 \newline 512x512 & 155K & ortho-rectified + \newline Manually registration\\ 
  \bottomrule
  \end{tabular}}
\end{table}
\subsection{Multimodal image matching method}
Existing optical-SAR matching methods can be categorized into three main types: traditional area-based methods, feature-based methods, and the recently popular deep learning-based methods.\cite{BaiZhu} Comprehensive reviews \cite{BaiZhu, JiayiMa, XingyuJiang} on this topic are available, and we suggest readers refer to these literature sources for more information and further exploration.

Feature-based methods extract handcrafted features from images and then employ key-point matching algorithms such as SIFT\cite{SIFT}, SURF\cite{SURF}, DAISY\cite{DAISY}, and others to perform the matching. While these feature-based methods have been successful in unimodal image registration, they can not extract a significant number of repetitive features from optical and SAR images due to significant geometric distortions and the nonlinear radiometric differences (NRD) \cite{BaiZhu, XingyuJiang}. In recent years, numerous efforts have been made to overcome the scale, rotation, radiance, and noise changes of MRSIs. Some researchers focus on designing more salient features instead of gradients to obtain more stable feature description. These distinctive features include OS-SIFT\cite{OS-SIFT}, RIFT\cite{RIFT}, MABPC\cite{NDHL}, HLMO\cite{MS-HLMO} and MSTM\cite{MSTM}. Besides, some methods use specially designed filter to reduces the NRD between multimodal images. For example, Yao et.al\cite{COFSM} use the co-occurrence filter to bulid a new scale space forfeature points extraction. Li et.al \cite{LNIFT} apply the local normalization filter to convert original images into normalized images for feature detection and description.

On the other hand, area-based methods involve mapping the template image to the corresponding window of the reference image and assessing similarity metrics between different image blocks. Commonly used similarity metrics include sum of squared differences (SSD)\cite{XingyuJiang}, normalized cross-correlation (NCC) \cite{NCC}, and mutual information(MI) \cite{MI}. To enhance the MMIM accuracy of area-based methods, HOPC\cite{HOPC} and OS-PC\cite{OS-PC} used phase congruency to describe the consistent scene structure between multi-modal images, while MIND\cite{MIND}, DLSS\cite{DLSS} and OMIRD\cite{OMIRD} aim to use the local self-similarity descriptors for template matching. Although these area-based approaches are commonly lack of scale and rotation invariance, they can be used to estimate the translation offsets exist in multimodal image images as the scaling and rotation transformations always can be corrected preliminarily according to the geo-referenced information acquired from GPS and IMU\cite{ZhangHan}.

In recent years, Deep learning(DL) has demonstrated great potential in image matching tasks. much of the recent research has focused on designing Convolutional Neural Networks (CNNs) for both detection\cite{r2d2, Dusmanu2019CVPR, detone18superpoint} and description\cite{HardNet, sosnet2019cvpr}. Trained with challenging data, CNNs largely improve the accuracy and robustness of matching. Besides, some research used transformers to design feature matching networks such as the LOFTR\cite{sun2021loftr}, lightGlue \cite{lightglue}, and ASpanFormer\cite{chen2022aspanformer}. However, these DL-based matching models are always trained on the image datasets without significant modality invariance like HPatches and MegaDepth. There are also some models specially designed for Optical-SAR image matching, for instance, Some models use the learning-based feature descriptor to obtain high-level features\cite{RDFM, MICM} In addition, some research aims to achieve image matching through end-to-end manner \cite{ArmandZampieri, ZhangHan, LloydHaydnHughes, MU-Net, E2EIR, RHWF, PRISE}. Furthermore, there are studies that introduce Generative Adversarial Networks(GAN) into the image matching process, thereby increasing the diversity of the training data and/or transforming images from different sources into similar-source images to reduce modality differences \cite{Merkle, Luca}.
\section{3MOS Construction}
\begin{figure}[h]
  \centering
  \includegraphics[height=4cm]{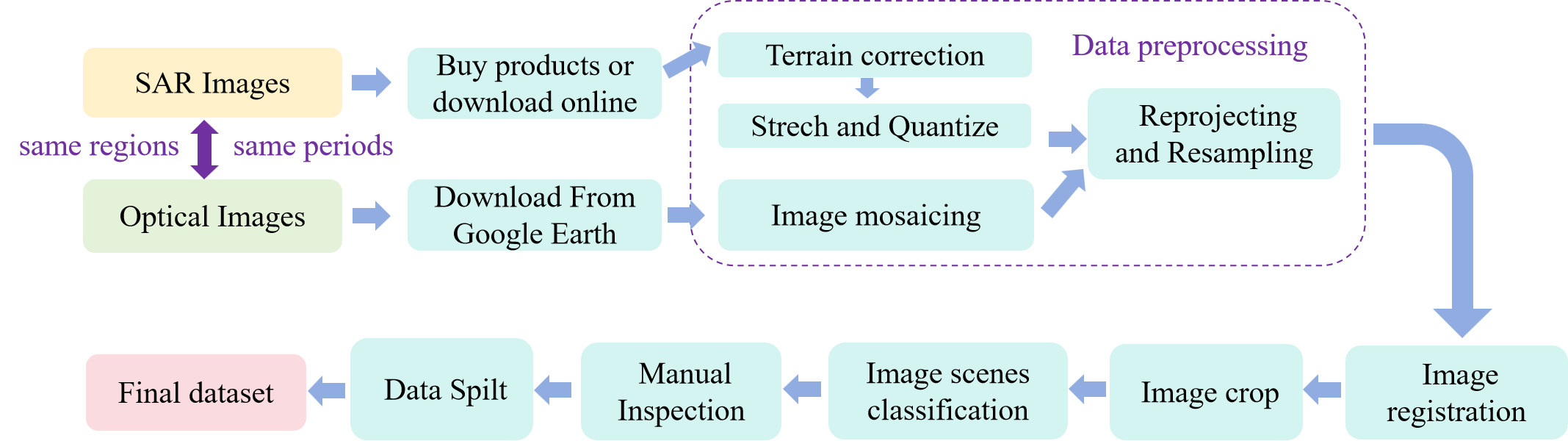}
  \caption{Workflow of 3MOS dataset construction procedure.
  }
  \label{fig:dataset_build}
\end{figure}
We used a total of 14 pairs of image data from 6 SAR satellites and Google Earth to construct the dataset. The details of these original image data can be found in appendix ~\ref{appendix:appendix1}, including the image geo-referenced location, imaging modes, imaging time, polarization method, image size etc.. The process of dataset construction is illustrated in \cref{fig:dataset_build} and can be divided into following 5 steps.
\subsection{Data Collection and  Preprocessing}
The original SAR data is either purchased by us or downloaded through public links provided by satellite companies, while the optical data is downloaded from historical imagery data in Google Earth and stitched together, ensuring that the coverage region and imaging time of the optical images match those of the SAR images.

For SAR images, we first use the openly available Shuttle Radar Topography Mission (SRTM) Digital Elevation Model (DEM) for terrain correction of the SAR images. Then, we apply grayscale stretching to enhance the images and quantize them to 8 bits. The optical images provided by Google Earth have already undergone georeferencing and quantization. Then, we reproject the images onto the WGS84 UTM coordinate system and resample them to the same spatial resolution.
\subsection{Image Registration}
\begin{figure}[t]
  \centering
  \includegraphics[height=7cm]{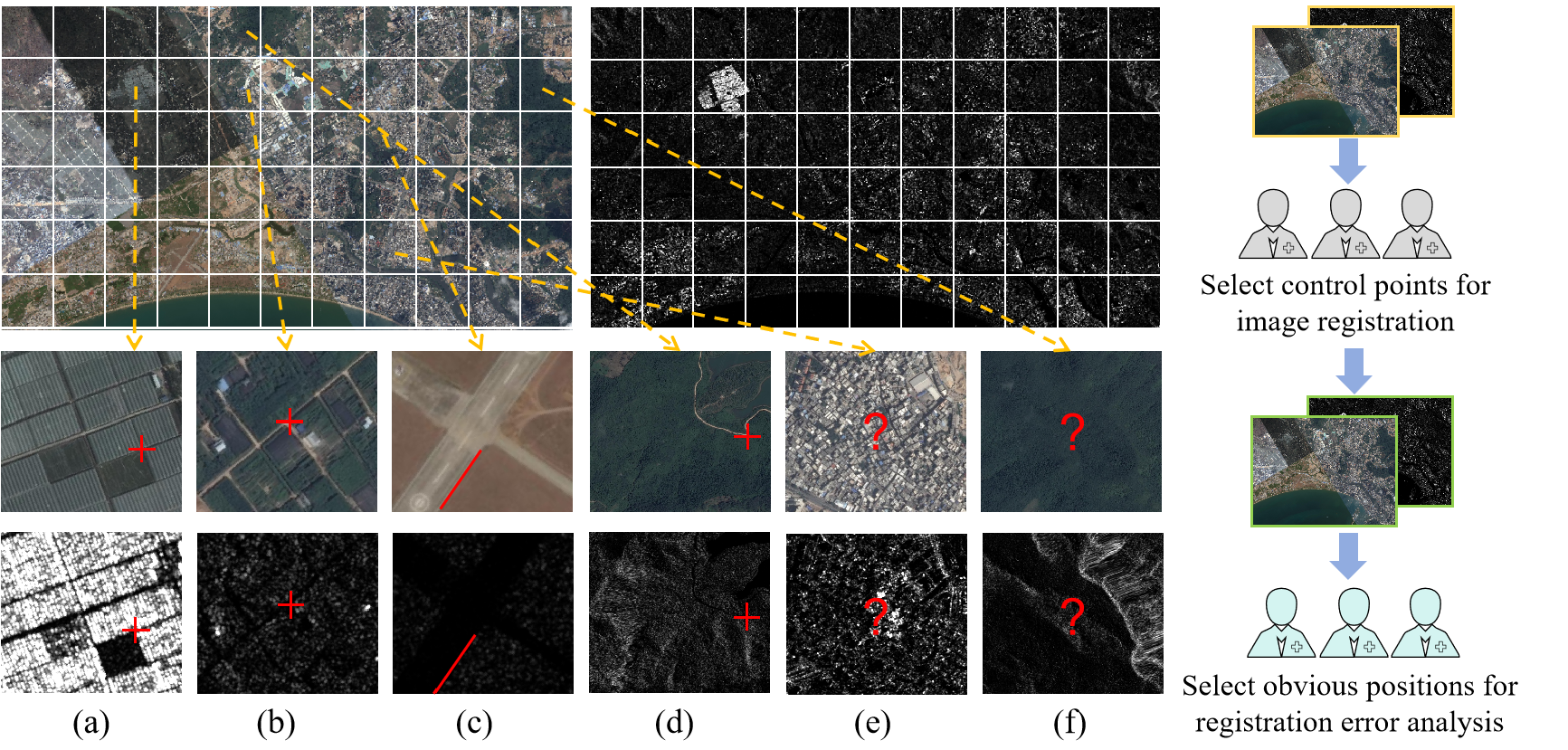}
  \caption{Selected control points for image registration and inspect the registration error manually.
  }
  \label{fig:register}
\end{figure}

Since the original optical and SAR images are from different data sources, there might be some geometric transformations between the preprocessed images of the same area, therefore image registration process is crucial. Referring to the image registration methods used in the current datasets(\cref{tab:datasets}), we argue that although using existing algorithms\cite{SOpatch, osdataset} for automatic registration is quick and relatively accurate, the registration performance may be unsatisfied for regions with severe noise interference or geometric feature differences. Additionally, registering data using algorithms introduces inherent errors related to the characteristics of the algorithm itself. This error will be learned by deep neural network, restricting the model's registration accuracy (which cannot exceed the registration method used during dataset preparation). To ensure registration accuracy and avoiding errors influenced by algorithms, we still choose to register the images manually. For each pair of optical and SAR images, we employed experienced experts to uniformly select dozens of control points with similar geometric structures and minimal noise, like the corner points in \cref{fig:register}(a)(b). Then, we employ a first-order polynomial transformation for image registration. 

Although we strive to enhance the registration accuracy, there still exists slight registration error within the dataset. To offer dataset users a qualitative grasp of the registration errors and to ensure minimal errors in the testing set, we invite other experts, not involved in the registration process, to assess the matching errors. The assessment method is similar to the matching procedure, involving the uniform selection of obvious positions to judge errors from the warped image pairs, such as corners and lines(\cref{fig:register}(c)(d)).
Experts were asked to categorize the positions into four classes based on error threshold: very high accuracy (error within 1 pixel, such as \cref{fig:register}(c)), high accuracy (error between 1 to 3 pixels,  such as \cref{fig:register}(d)), low accuracy (error above 3 pixels), and indiscernible regions(\cref{fig:register}(e)(f)). After that, we remove the indiscernible regions and calculate the matching error of different satellites data, as shown in \cref{tab:error}.

\begin{table}[h]
  \caption{Registration error distribution of different satellites' data.
  }
  \label{tab:error}
  \centering
  \scalebox{0.6}{
  \begin{tabular}{ccc} 
  \toprule
\textbf{Satellite} & \textbf{Mean error(pixel)} & \textbf{Standard deviation(pixel)}  \\
\midrule
SEN1 & 0.31 & 0.34  \\
TerraSAR & 0.52 & 0.63  \\
GF3 & 0.81 & 0.71  \\
Radarsat & 0.74 & 0.66  \\
ALOS & 0.79 & 0.81  \\
RCM & 1.20 & 1.1  \\
  \bottomrule
  \end{tabular}}
  \end{table}
  
\subsection{Image Crop}
Due to the large size of the original images, they need to be divided into smaller blocks for model training and testing. The size of image blocks are determined on the spatial distance. Specifically, when the spatial distance is 1.25m, the image size is set to 512x512. When the  spatial distance is greater than or equal to 2.5m, the image size is set to 256x256. This selection aims to ensure that each image block covers a sufficiently large area to provide more texture information while avoiding excessively large image sizes that could affect subsequent training and testing efficiency. We set the overlap rate of the images to 50\%\cite{SEN12,SEN12MS}, allowing us to generate a larger number of image blocks while maintaining content diversity.
\subsection{Image Scene Classification}
In remote sensing images, different scenes exhibit distinct characteristics. For example, SAR images of urban areas typically show strong backscatter due to man-made structures, and they often display a high level of clutter and speckle noise, which can make feature identification challenging. On the other hand, SAR images of desert areas are usually characterized by low backscatter because of the sparse vegetation and lack of moisture in the terrain. Unique geological features such as sand patterns and rock formations can be highlighted in SAR images of deserts due to their surface roughness.

\begin{figure}[h]
  \centering
  \includegraphics[height=4.5cm]{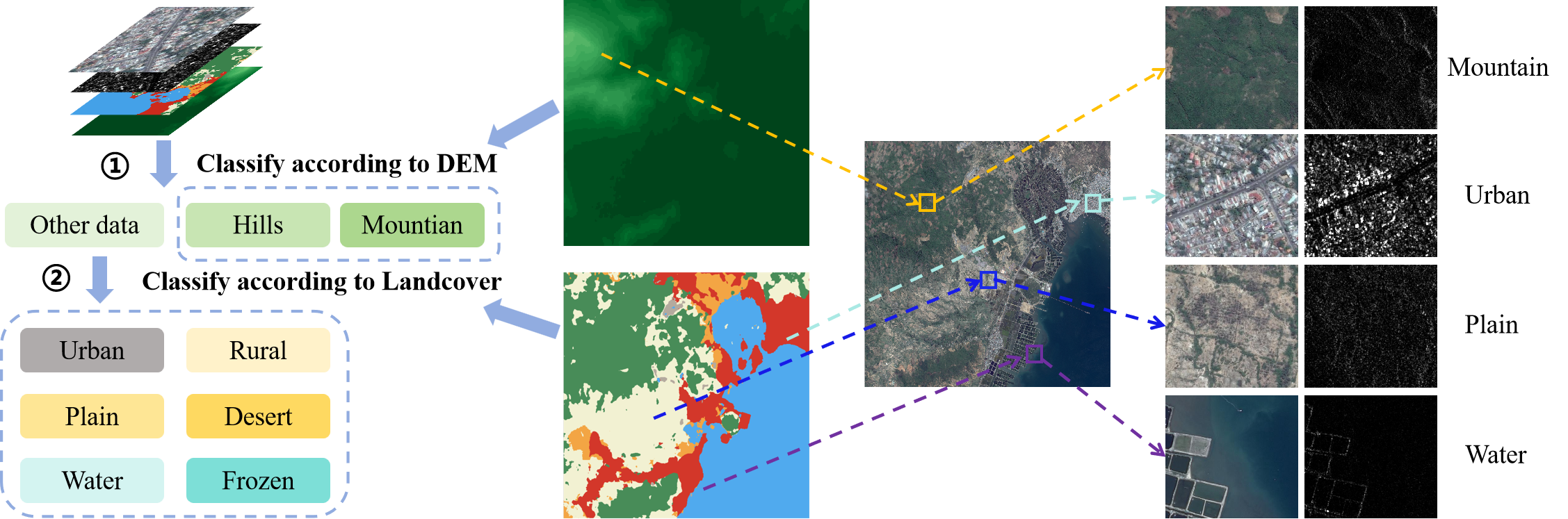}
  \caption{The flowchart of image scene classification.
  }
  \label{fig:classification}
\end{figure}

Furthermore, the difficulty of image matching varies across different scenes. Generally, matching SAR images of desert and mountainous areas is more challenging compared to urban areas. A reliable matching algorithm should have robust performance across various scenes. Therefore, we classified the images according to different ground scenes. Specifically, we utilized NASADEM data\footnote{\url{https://e4ftl01.cr.usgs.gov/DP109/MEASURES/NASADEM_HGT.001/2000.02.11/}} and Sentinel-2 Landcover data\footnote{\url{https://www.arcgis.com/apps/instant/media/index.html?appid=fc92d38533d440078f17678ebc20e8e2}} to automatically classify the image scenes and obtained eight categories of image data, including urban, rural, plains, hills, mountains, water, desert, and frozen earth. As shown in \cref{fig:classification}, we use the DEM data to perform the first classification based on the elevation differences within the image area. For example, areas with elevation differences greater than 150m are classified as mountains, while areas with elevation differences between 50m and 150m are classified as hills. For areas with elevation differences less than 50m, we use landcover data to perform a second classification, primarily based on which type of landcover dominates within the image. For example, if there are more buildings in an image, it is considered an urban area. If the buildings are sparse and there is more vegetation, it is considered a plain. For more information on scene characteristics and scene classification details, please refer to appendix ~\ref{appendix:appendix2}.

\subsection{Manual Inspection and Data Spilt}
Due to the presence of invalid areas in the segmented dataset, such as large no-data areas, regions with high cloud coverage (over 30\%), completely featureless areas (e.g., open sea), and areas with imaging quality issues (e.g., severely distorted colors or spatial resolution), it is not meaningful to perform image matching on these data. Therefore, we employ manual inspection to remove these regions and enhance the overall quality of the data.
\begin{figure}[h]
  \centering
  \includegraphics[height=3cm]{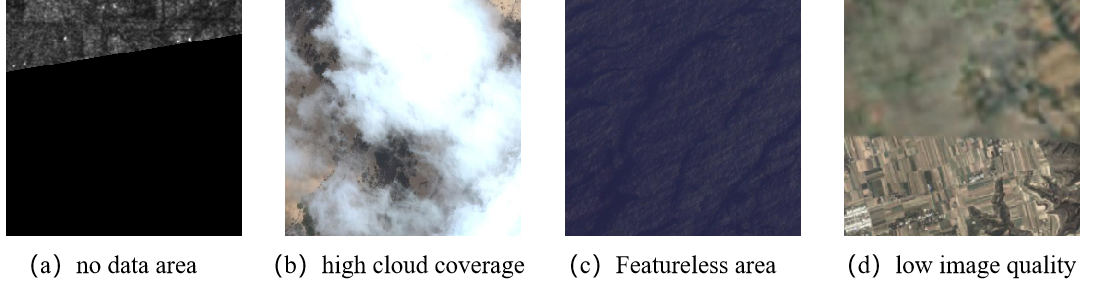}
  \caption{The useless images have been deleted.
  }
  \label{fig:useless}
\end{figure}

For data from different satellites and different scenes, we randomly spilt the dataset into training, validation, and testing sets at a ratio of 6:2:2. This is done to evaluate the model’s matching performance across different satellites, resolutions, and scenes.

\subsection{Data Statistics}
\begin{figure}[h]
  \centering
  \includegraphics[height=4cm]{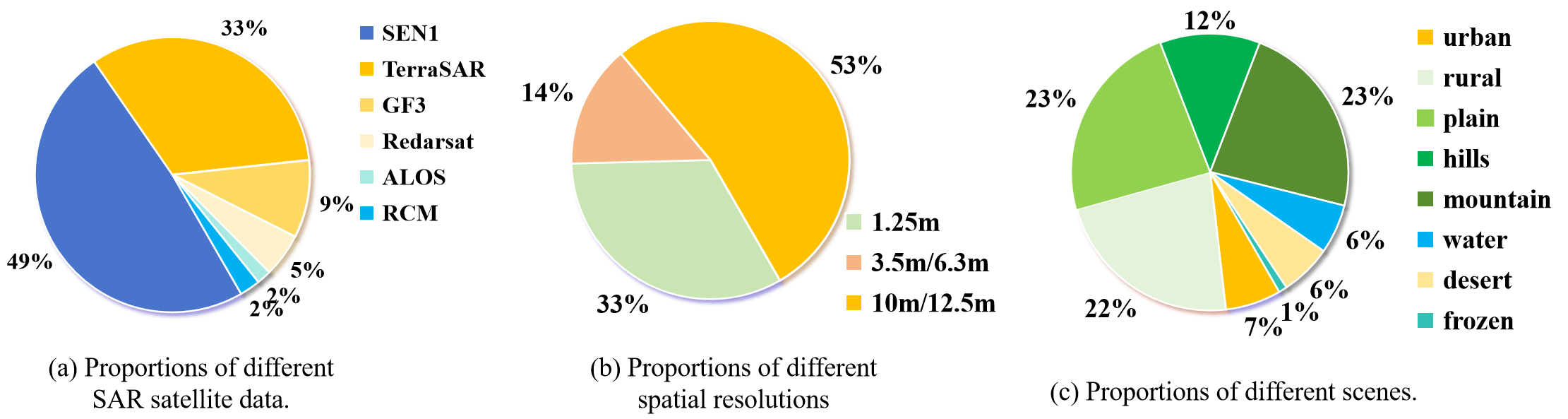}
  \caption{Pie chart of data distribution for different sources, resolutions, and scenes.
  }
  \label{fig:data_pie}
\end{figure}

The statistical analysis of 3MOS dataset are presented in \cref{tab:stastics} and \cref{fig:data_pie}, there are SAR data from multiple sources including GF3, TerraSAR, ALOS, Radarsat, SEN1, and RCM(RadarSat Constellation Mission), with various spatial resolutions ranging from 1.25 to 12.5 meters. The dataset has eight different scenes from multiple regions around the world. it shuoud be note that SEN1 and TerraSAR are the main data sources, with data from other satellites making up 18\% due to their greater difficulty in acquisition. In addition, there is variation in the proportions of different scenes, primarily influenced by satellite coverage. Many large water areas have been excluded, and data for frozen earth and desert scenes are particularly scarce and difficult to register.
\begin{table}[h]
  \caption{Data statistics of 3MOS. SR:spatial resolution. GE: Google earth.
  }
  \label{tab:stastics}
  \centering
  \scalebox{0.7}{
  \begin{tabular}{lcccll} 
  \toprule
\textbf{SAR} & \textbf{Optical} & \textbf{SR(m)} & \textbf{Num.} & \textbf{Image scenes} & \textbf{Location} \\
\midrule
TerraSAR & GE & 1.25 & 51195 & urban, rural, plain, hills, mountain, water & hainan in China and Vietnam \\
GF3 & GE & 3.5 & 14209 & urban, rural, plain, water & hunan in China \\
Radarsat & GE & 6.3 & 7958 & urban, rural, hills, mountain & shanxi in China \\
SEN1 & GE & 10 & 75720 & urban, rural, plain, hills, mountain, water, desert, frozen & \begin{tabular}[c]{@{}l@{}}China, the USA, Netherlands, Iraq, Zimbabwe\\ ,Antarctica\end{tabular} \\
ALOS & GE & 12.5 & 2650 & desert & gansu in China \\
RCM & GE & 12.5 & 3745 & urban, rural, plain & sicuan in China \\

  \bottomrule
  \end{tabular}}
\end{table}

\section{Evaluation}

\begin{figure}[h]
  \centering
  \includegraphics[height=3cm]{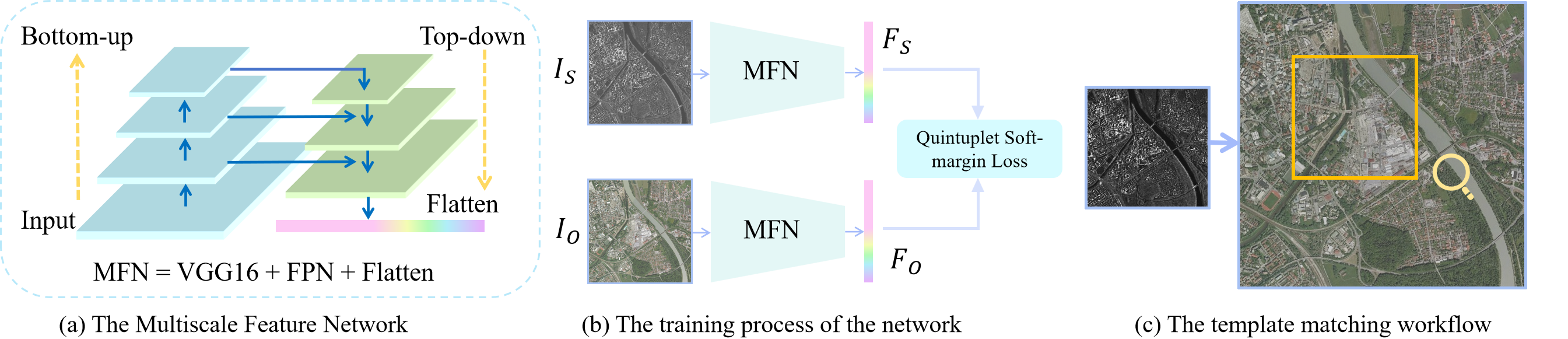}
  \caption{The multiscale feature network using in the template matching experiment.
  }
  \label{fig:MFN}
  
\end{figure}

In this section, we describe our experimental setups and the implementation details of baseline methods. We conducted template matching experiments on the 3MOS dataset using six traditional multimodal image matching methods and one deep learning method. Template matching involves locating the position of an NxN template image in an MxM reference image (\cref{fig:MFN}(c)), where mainly translation transformations exist between the reference and template images.
 
The correct matching ratio(CMR) is used as evaluation metric. 
${CMR} = \frac{{{N_T}}}{N} \times 100\% $, where N is the total number of test image pairs and $N_T$ represents the number of test image pairs with errors (Euclidean distance) less than the threshold $T$. The multi-scale feature network(MFN) used in the experiment was proposed by us, drawing inspiration from FPN\cite{MaskR-CNN, FasterR-CNN} and contrastive learning\cite{YifanSun,XunWang}. The network architecture, as shown in \cref{fig:MFN}, initially extracts multi-scale image features using VGG16+FPN (\cref{fig:MFN}(a)), followed by flattening the features and employing a quintuplet loss for contrastive learning (\cref{fig:MFN}(b)). the specific details on loss function design and network training are discussed in the supplementary materials.

\subsection{Template matching on different data}
We first conducted matching experiments on images from different satellites and scenes, the experimental results are summarized in \cref{tab:result2}.
For the same type of SAR satellite data, we train the MFN using data from all scenes. The template image size is half of the reference image size, and the template image position (ground truth) is randomly selected on the reference image. All methods are tested on the same number of test sets and the same ground truth.

From \cref{tab:result2}, we can observe the following points. Firstly, we find none of these methods achieve consistently superior performance across different sources, resolutions and scenes. Besides, the matching capabilities of the same method vary significantly under different satellites and resolutions. For instance, the matching results of urban and water scenes in the SEN dataset are noticeably better than other satellite data of the same scenes (within 10 pixels of CMR). This may be attributed to the lower resolution of the SEN1 dataset and its superior imaging quality, making it easier to match. 

Secondly, the matching performance of the same method varies significantly under different scenes of the same satellite. For example, in the SEN1 dataset, the CMR for urban scenes is much higher than that for water, desert, and frozen earth scenes. This is mainly because urban scenes have more artificial structures with distinct features, while the other three scenes have fewer effective features. For instance, the optical images in the frozen earth region are covered by ice and snow, while SAR images penetrate through them, leading to significant modal differences and increased matching difficulty. Additionally, we found that our designed MFN architecture achieved state-of-the-art results on TerraSAR, GF3, SEN1, and Radarsat datasets. This demonstrates the superiority of deep learning over traditional methods when the data sources are the same, especially in high-resolution TerraSAR datasets.

\begin{table}[h]
  \caption{Template matching result of different data sources and different scenes. The evaluation metric is CMR(\%). mtns: mountains.}
  \label{tab:result2}
  \centering
  \begin{subtable}{0.22\linewidth}
        \caption{GF3 urban}
  \label{tab:GF3-urban}
  \centering
  \scalebox{0.6}{
  \begin{tabular}{ccccc}
    \hline
     Method & $\leq 3$ & $\leq 6$ & $\leq 10$ & $\leq 20$ \\
    \hline
    MFN & \textbf{47.2} & \textbf{77.8} & \textbf{86.1} & \textbf{95.0} \\
    HOPC\cite{HOPC} & 40.6 & 68.3 & 82.8 & 86.7 \\
    CFOG\cite{CFOG} & 38.9 & 62.8 & 76.1 & 81.7 \\
    NCC\cite{NCC} & 6.7 & 15.0 & 21.1 & 30.0 \\
    NMI\cite{MI} & 19.4 & 33.3 & 42.8 & 50.0 \\
    MIND\cite{MIND} & 23.9 & 43.9 & 57.8 & 67.2 \\
    OMIRD\cite{OMIRD} & 18.3 & 37.8 & 50.6 & 55.0 \\
    \hline
  \end{tabular}}
  \end{subtable}
  \hfill
  \begin{subtable}{0.22\linewidth}
        \caption{GF3 plain}
  \label{tab:GF3-plain}
  \centering
  \scalebox{0.6}{
  \begin{tabular}{ccccc}
    \hline
     Method & $\leq 3$ & $\leq 6$ & $\leq 10$ & $\leq 20$ \\
    \hline
MFN &\textbf{ 43.9} &\textbf{ 69.2} & \textbf{78.9} & \textbf{83.2} \\
HOPC & 23.8 & 57.6 & 72.4 & 78.2 \\
CFOG & 22.8 & 53.4 & 69.7 & 74.4 \\
NCC & 7.8 & 15.3 & 27.1 & 37.8 \\
NMI & 13.8 & 32.6 & 48.6 & 60.9 \\
MIND & 17.8 & 43.1 & 58.1 & 63.2 \\
OMIRD & 16.5 & 36.8 & 47.9 & 57.1 \\
    \hline
  \end{tabular}}
  \end{subtable}
  \hfill
  \begin{subtable}{0.22\linewidth}
        \caption{GF3 rural}
  \label{tab:GF3-rural}
  \centering
  \scalebox{0.6}{
  \begin{tabular}{ccccc}
    \hline
     Method & $\leq 3$ & $\leq 6$ & $\leq 10$ & $\leq 20$ \\
    \hline
MFN & \textbf{59.0} & \textbf{85.6} & \textbf{91.3} & \textbf{94.7} \\
HOPC & 40.5 & 66.1 & 75.8 & 79.7 \\
CFOG & 37.7 & 62.8 & 72.9 & 77.7 \\
NCC & 10.0 & 20.3 & 27.4 & 35.9 \\
NMI & 22.0 & 43.1 & 52.3 & 60.3 \\
MIND & 27.9 & 49.0 & 58.0 & 63.9 \\
OMIRD & 20.4 & 37.8 & 45.7 & 53.3 \\
    \hline
  \end{tabular}}
  \end{subtable}
  \hfill
  \begin{subtable}{0.22\linewidth}
        \caption{GF3 water}
  \label{tab:GF3-water}
  \centering
  \scalebox{0.6}{
  \begin{tabular}{ccccc}
    \hline
     Method & $\leq 3$ & $\leq 6$ & $\leq 10$ & $\leq 20$ \\
    \hline
MFN & \textbf{56.4} & \textbf{79.1} & \textbf{84.2} & \textbf{88.7} \\
HOPC & 35.6 & 70.2 & 80.2 & 85.0 \\
CFOG & 29.6 & 64.1 & 79.7 & 85.3 \\
NCC & 8.1 & 17.1 & 23.8 & 31.1 \\
NMI & 19.0 & 40.3 & 49.0 & 56.2 \\
MIND & 26.2 & 56.5 & 70.2 & 77.5 \\
OMIRD & 22.9 & 46.2 & 55.7 & 63.8 \\
    \hline
  \end{tabular}}
  \end{subtable}
  \hfill
   \begin{subtable}{0.22\linewidth}
        \caption{Radarsat urban}
  \label{tab:Radarsat-city}
  \centering
  \scalebox{0.6}{
  \begin{tabular}{ccccc}
    \hline
     Method & $\leq 3$ & $\leq 6$ & $\leq 10$ & $\leq 20$ \\
    \hline
MFN & \textbf{95.8} & \textbf{98.1} & \textbf{98.1} & \textbf{98.6} \\
HOPC & 92.6 & 96.8 & 96.8 & 96.8 \\
CFOG & 86.1 & 94.9 & 94.9 & 94.9 \\
NCC & 26.9 & 44.4 & 50.5 & 57.9 \\
NMI & 27.8 & 38.9 & 41.2 & 44.9 \\
MIND & 72.2 & 79.2 & 81.5 & 82.9 \\
OMIRD & 68.1 & 81.0 & 83.3 & 86.1 \\
    \hline
  \end{tabular}}
  \end{subtable}
  \hfill
  \begin{subtable}{0.22\linewidth}
        \caption{Radarsat hills}
  \label{tab:Radarsat-hill}
  \centering
  \scalebox{0.6}{
  \begin{tabular}{ccccc}
    \hline
     Method & $\leq 3$ & $\leq 6$ & $\leq 10$ & $\leq 20$ \\
    \hline
MFN & \textbf{89.8 }& \textbf{97.3} & \textbf{97.7} & \textbf{98.1} \\
HOPC & 80.3 & 86.0 & 87.5 & 89.0 \\
CFOG & 61.0 & 69.7 & 72.3 & 75.4 \\
NCC & 27.7 & 37.1 & 41.7 & 47.7 \\
NMI & 35.2 & 44.3 & 48.1 & 49.2 \\
MIND & 60.2 & 71.2 & 74.2 & 76.1 \\
OMIRD & 51.1 & 61.7 & 64.8 & 67.4 \\
    \hline
  \end{tabular}}
  \end{subtable}
  \hfill
  \begin{subtable}{0.22\linewidth}
        \caption{Radarsat rural}
  \label{tab:Radarsat-rural}
  \centering
  \scalebox{0.6}{
  \begin{tabular}{ccccc}
    \hline
     Method & $\leq 3$ & $\leq 6$ & $\leq 10$ & $\leq 20$ \\
    \hline
MFN & \textbf{95.8} & \textbf{98.1} & \textbf{98.1} & \textbf{98.6} \\
HOPC & 92.6 & 96.8 & 96.8 & 96.8 \\
CFOG & 86.1 & 94.9 & 94.9 & 94.9 \\
NCC & 26.9 & 44.4 & 50.5 & 57.9 \\
NMI & 27.8 & 38.9 & 41.2 & 44.9 \\
MIND & 72.2 & 79.2 & 81.5 & 82.9 \\
OMIRD & 68.1 & 81.0 & 83.3 & 86.1 \\
    \hline
  \end{tabular}}
  \end{subtable}
  \hfill
  \begin{subtable}{0.22\linewidth}
        \caption{Radarsat mtns}
  \label{tab:Radarsat-mountains}
  \centering
  \scalebox{0.6}{
  \begin{tabular}{ccccc}
    \hline
     Method & $\leq 3$ & $\leq 6$ & $\leq 10$ & $\leq 20$ \\
    \hline
    MFN & \textbf{51.4} & \textbf{73.2} & \textbf{82.3} &\textbf{ 90.0} \\
    HOPC & 41.4 & 59.5 & 68.6 & 77.7 \\
    CFOG & 35.5 & 54.5 & 65.5 & 73.2 \\
    NCC & 3.6 & 6.4 & 10.9 & 18.2 \\
    NMI & 12.3 & 20.5 & 23.6 & 30.9 \\
    MIND & 28.6 & 45.5 & 58.2 & 69.5 \\
    OMIRD & 26.4 & 43.6 & 56.4 & 64.5 \\
        \hline
  \end{tabular}}
  \end{subtable}
  \hfill
   \begin{subtable}{0.22\linewidth}
        \caption{RCM rural}
  \label{tab:RCM rural}
  \centering
  \scalebox{0.6}{
  \begin{tabular}{ccccc}
    \hline
     Method & $\leq 3$ & $\leq 6$ & $\leq 10$ & $\leq 20$ \\
    \hline
MFN & 15.3 & 38.6 & 50.3 & 56.9 \\
HOPC & \textbf{28.4} &\textbf{ 54.2} & \textbf{71.6} & \textbf{78.4} \\
CFOG & 20.4 & 46.7 & 59.3 & 65.9 \\
NCC & 12.0 & 24.9 & 38.0 & 47.3 \\
NMI & 8.1 & 16.8 & 24.3 & 28.7 \\
MIND & 19.8 & 40.4 & 52.7 & 59.3 \\
OMIRD & 14.7 & 29.6 & 39.2 & 45.2 \\
    \hline
  \end{tabular}}
  \end{subtable}
  \hfill
  \begin{subtable}{0.22\linewidth}
        \caption{RCM urban}
  \label{tab:RCM urban}
  \centering
  \scalebox{0.6}{
  \begin{tabular}{ccccc}
    \hline
     Method & $\leq 3$ & $\leq 6$ & $\leq 10$ & $\leq 20$ \\
    \hline
MFN & 24.3 & 46.8 & 52.5 & 55.1 \\
HOPC & \textbf{55.5} & \textbf{80.7} & \textbf{88.7 }&\textbf{ 91.0} \\
CFOG & 48.8 & 74.1 & 81.7 & 84.7 \\
NCC & 18.6 & 33.2 & 44.2 & 48.5 \\
NMI & 9.0 & 14.6 & 19.6 & 23.9 \\
MIND & 35.2 & 59.1 & 67.8 & 70.1 \\
OMIRD & 28.2 & 50.5 & 57.8 & 61.8 \\
    \hline
  \end{tabular}}
  \end{subtable}
  \hfill
  \begin{subtable}{0.22\linewidth}
        \caption{RCM plain}
  \label{tab:RCM plain}
  \centering
  \scalebox{0.6}{
  \begin{tabular}{ccccc}
    \hline
     Method & $\leq 3$ & $\leq 6$ & $\leq 10$ & $\leq 20$ \\
    \hline
MFN & 3.4 & 19.7 & 30.8 & 37.6 \\
HOPC & \textbf{14.5 }& \textbf{43.6} & \textbf{65.0} & \textbf{69.2} \\
CFOG & 10.3 & 34.2 & 54.7 & 60.7 \\
NCC & 7.7 & 22.2 & 40.2 & 51.3 \\
NMI & 3.4 & 10.3 & 14.5 & 18.8 \\
MIND & 14.5 & 33.3 & 49.6 & 53.0 \\
OMIRD & 6.0 & 18.8 & 26.5 & 31.6 \\
    \hline
  \end{tabular}}
  \end{subtable}
  \hfill
  \begin{subtable}{0.22\linewidth}
        \caption{ALOS desert}
  \label{tab:ALOS desert}
  \centering
  \scalebox{0.6}{
  \begin{tabular}{ccccc}
    \hline
     Method & $\leq 3$ & $\leq 6$ & $\leq 10$ & $\leq 20$ \\
    \hline
MFN & 27.7 & 31.5 & 32.3 & 36.2 \\
HOPC & \textbf{33.4} & \textbf{37.0} & \textbf{38.3} & \textbf{40.9} \\
CFOG & 20.0 & 21.7 & 22.6 & 26.4 \\
NCC & 4.2 & 9.6 & 11.9 & 15.3 \\
NMI & 15.7 & 19.4 & 20.9 & 24.9 \\
MIND & 33.4 & 36.6 & 37.9 & 40.6 \\
OMIRD & 21.1 & 24.7 & 26.2 & 29.6 \\
        \hline
  \end{tabular}}
  \end{subtable}

    \hfill
   \begin{subtable}{0.22\linewidth}
        \caption{SEN1 frozen}
  \label{SEN frozen}
  \centering
  \scalebox{0.6}{
  \begin{tabular}{ccccc}
    \hline
     Method & $\leq 3$ & $\leq 6$ & $\leq 10$ & $\leq 20$ \\
    \hline
MFN & \textbf{16.3} &\textbf{ 37.0} & \textbf{47.3 }& \textbf{62.3} \\
HOPC & 5.3 & 12.7 & 17.0 & 24.3 \\
CFOG & 4.0 & 8.3 & 13.0 & 20.7 \\
NCC & 2.0 & 7.0 & 10.7 & 20.0 \\
NMI & 2.0 & 4.3 & 12.0 & 19.3 \\
MIND & 4.3 & 7.7 & 14.3 & 20.3 \\
OMIRD & 2.0 & 6.7 & 10.7 & 19.3 \\
    \hline
  \end{tabular}}
  \end{subtable}
  \hfill
  \begin{subtable}{0.22\linewidth}
        \caption{SEN1 desert}
  \label{tab:SEN desert}
  \centering
  \scalebox{0.6}{
  \begin{tabular}{ccccc}
    \hline
     Method & $\leq 3$ & $\leq 6$ & $\leq 10$ & $\leq 20$ \\
    \hline
MFN & \textbf{42.0} & \textbf{49.5} & \textbf{52.0} & \textbf{56.2} \\
HOPC & 19.1 & 21.9 & 23.6 & 28.1 \\
CFOG & 13.3 & 15.0 & 16.5 & 22.4 \\
NCC & 2.8 & 3.6 & 4.5 & 7.5 \\
NMI & 11.1 & 13.1 & 14.4 & 17.6 \\
MIND & 22.7 & 24.8 & 27.0 & 31.9 \\
OMIRD & 14.1 & 17.3 & 19.1 & 24.5 \\
    \hline
  \end{tabular}}
  \end{subtable}
  \hfill
  \begin{subtable}{0.22\linewidth}
        \caption{SEN1 urban}
  \label{tab:SEN urban}
  \centering
  \scalebox{0.6}{
  \begin{tabular}{ccccc}
    \hline
     Method & $\leq 3$ & $\leq 6$ & $\leq 10$ & $\leq 20$ \\
    \hline
MFN & \textbf{95.8} & \textbf{98.1} & \textbf{98.1} & \textbf{98.6} \\
HOPC & 92.6 & 96.8 & 96.8 & 96.8 \\
CFOG & 86.1 & 94.9 & 94.9 & 94.9 \\
NCC & 26.9 & 44.4 & 50.5 & 57.9 \\
NMI & 27.8 & 38.9 & 41.2 & 44.9 \\
MIND & 72.2 & 79.2 & 81.5 & 82.9 \\
OMIRD & 68.1 & 81.0 & 83.3 & 86.1 \\
    \hline
  \end{tabular}}
  \end{subtable}
  \hfill
  \begin{subtable}{0.22\linewidth}
        \caption{SEN1 water}
  \label{tab:SEN water}
  \centering
  \scalebox{0.6}{
  \begin{tabular}{ccccc}
    \hline
     Method & $\leq 3$ & $\leq 6$ & $\leq 10$ & $\leq 20$ \\
    \hline
MFN & \textbf{79.5} & \textbf{84.6} & \textbf{85.2} & \textbf{86.9} \\
HOPC & 73.8 & 77.4 & 78.6 & 80.8 \\
CFOG & 66.8 & 70.2 & 71.5 & 73.6 \\
NCC & 19.2 & 22.6 & 26.8 & 31.9 \\
NMI & 37.4 & 44.8 & 48.4 & 53.1 \\
MIND & 66.4 & 69.1 & 71.0 & 74.2 \\
OMIRD & 54.3 & 59.2 & 61.7 & 65.3 \\
        \hline
  \end{tabular}}
  \end{subtable}

    \hfill
   \begin{subtable}{0.22\linewidth}
        \caption{TerraSAR mtns}
  \label{tab:TerraSAR mtns}
  \centering
  \scalebox{0.6}{
  \begin{tabular}{ccccc}
    \hline
     Method & $\leq 3$ & $\leq 6$ & $\leq 10$ & $\leq 20$ \\
    \hline
MFN & \textbf{11.0} & \textbf{28.2} & \textbf{42.0} & \textbf{54.6} \\
HOPC & 1.1 & 3.2 & 5.4 & 8.4 \\
CFOG & 0.8 & 2.3 & 3.7 & 5.5 \\
NCC & 1.1 & 3.4 & 6.5 & 12.3 \\
NMI & 1.2 & 3.0 & 6.0 & 10.3 \\
MIND & 0.7 & 2.1 & 3.8 & 6.2 \\
OMIRD & 0.7 & 1.8 & 3.2 & 6.0 \\
    \hline
  \end{tabular}}
  \end{subtable}
  \hfill
  \begin{subtable}{0.22\linewidth}
        \caption{TerraSAR plain}
  \label{tab:TerraSAR plain}
  \centering
  \scalebox{0.6}{
  \begin{tabular}{ccccc}
    \hline
     Method & $\leq 3$ & $\leq 6$ & $\leq 10$ & $\leq 20$ \\
    \hline
MFN & \textbf{52.3} & \textbf{87.0} & \textbf{95.1} & \textbf{97.0} \\
HOPC & 19.2 & 36.5 & 42.5 & 46.9 \\
CFOG & 11.8 & 24.0 & 28.7 & 31.7 \\
NCC & 0.8 & 2.4 & 5.7 & 10.5 \\
NMI & 18.2 & 33.5 & 41.1 & 44.0 \\
MIND & 14.9 & 26.7 & 31.9 & 35.7 \\
OMIRD & 14.3 & 24.0 & 29.2 & 33.4 \\
    \hline
  \end{tabular}}
  \end{subtable}
  \hfill
  \begin{subtable}{0.22\linewidth}
        \caption{TerraSAR urban}
  \label{tab:TerraSAR urban}
  \centering
  \scalebox{0.6}{
  \begin{tabular}{ccccc}
    \hline
     Method & $\leq 3$ & $\leq 6$ & $\leq 10$ & $\leq 20$ \\
    \hline
MFN & \textbf{47.7} & \textbf{86.1} & \textbf{96.6} & \textbf{99.0} \\
HOPC & 26.1 & 59.1 & 76.7 & 80.4 \\
CFOG & 21.2 & 43.4 & 59.7 & 65.4 \\
NCC & 3.6 & 9.1 & 14.9 & 19.9 \\
NMI & 19.4 & 39.9 & 51.1 & 56.0 \\
MIND & 15.8 & 38.1 & 51.9 & 57.4 \\
OMIRD & 13.4 & 31.9 & 43.8 & 49.1 \\
    \hline
  \end{tabular}}
  \end{subtable}
  \hfill
  \begin{subtable}{0.22\linewidth}
        \caption{TerraSAR water}
  \label{tab:TerraSAR water}
  \centering
  \scalebox{0.6}{
  \begin{tabular}{ccccc}
    \hline
     Method & $\leq 3$ & $\leq 6$ & $\leq 10$ & $\leq 20$ \\
    \hline
MFN & \textbf{33.8} & \textbf{57.8} & \textbf{68.5} & \textbf{74.2} \\
HOPC & 17.7 & 34.9 & 43.9 & 50.6 \\
CFOG & 15.4 & 29.9 & 36.0 & 44.6 \\
NCC & 8.7 & 14.7 & 19.8 & 26.5 \\
NMI & 16.2 & 30.2 & 40.3 & 47.9 \\
MIND & 14.7 & 28.2 & 35.3 & 41.6 \\
OMIRD & 11.0 & 19.5 & 24.9 & 31.5 \\
        \hline
  \end{tabular}}
  \end{subtable}
\end{table}

\begin{table}[h]
\caption{Template matching result of MPN using different training data. The evaluation metric is CMR(\%). mtns: mountains. TS: TerraSAR. T. data: Training data.}
 \label{tab:result1}
  \centering
  \begin{subtable}{0.22\linewidth}
        \caption{Test: TS urban}
  \label{tab:Test set: TerraSAR water}
  \centering
  \scalebox{0.6}{
  \begin{tabular}{ccccc}
    \hline
     T. data & $\leq 3$ & $\leq 6$ & $\leq 10$ & $\leq 20$ \\
    \hline
    TerraSAR & \textbf{33.8} & \textbf{57.8} & \textbf{68.5} & \textbf{74.2} \\
    Radarsat & 20.1 & 44.9 & 56.0 & 66.1 \\
   RCM & 17.2 & 35.3 & 44.0 & 51.7 \\
    GF3 & 20.7 & 38.5 & 46.6 & 56.7 \\
    ALOS & 21.5 & 42.3 & 50.3 & 56.4 \\
    HOPC* & 17.7 & 34.9 & 43.9 & 50.6 \\
    \hline
  \end{tabular}}
  \end{subtable}
  \hfill
  \begin{subtable}{0.22\linewidth}
        \caption{Test: TS urban}
  \label{tab:Test set: TerraSAR urban}
  \centering
  \scalebox{0.6}{
  \begin{tabular}{ccccc}
    \hline
     T. data & $\leq 3$ & $\leq 6$ & $\leq 10$ & $\leq 20$ \\
    \hline
    TerraSAR & \textbf{47.7} & \textbf{86.1} & \textbf{96.6} & \textbf{99.0} \\
    Radarsat & 22.6 & 64.9 & 85.1 & 93.4 \\
   RCM & 11.0 & 36.4 & 49.1 & 55.1 \\
    GF3 & 11.4 & 27.8 & 35.3 & 39.8 \\
    ALOS & 20.0 & 53.2 & 69.4 & 75.6 \\
    HOPC* & 26.1 & 59.1 & 76.7 & 80.4 \\
    \hline
  \end{tabular}}
  \end{subtable}
  \hfill
  \begin{subtable}{0.22\linewidth}
        \caption{Test: SEN1 water}
  \label{tab:Test set: SEN water}
  \centering
  \scalebox{0.6}{
  \begin{tabular}{ccccc}
    \hline
     T. data & $\leq 3$ & $\leq 6$ & $\leq 10$ & $\leq 20$ \\
    \hline
    SEN1 &  \textbf{79.5} & \textbf{84.6} & \textbf{85.2} & \textbf{86.9} \\
    Radarsat & 64.7 & 75.5 & 77.0 & 80.1 \\
   RCM & 45.0 & 58.1 & 59.6 & 63.2 \\
    GF3 & 41.2 & 53.9 & 57.3 & 59.4 \\
    ALOS & 61.5 & 68.9 & 70.6 & 72.5 \\
    HOPC* & 73.8 & 77.4 & 78.6 & 80.8 \\
    \hline
  \end{tabular}}
  \end{subtable}
  \hfill
  \begin{subtable}{0.22\linewidth}
        \caption{Test: SEN1 urban}
  \label{tab:Test set: SEN urban}
  \centering
  \scalebox{0.6}{
  \begin{tabular}{ccccc}
    \hline
     T. data & $\leq 3$ & $\leq 6$ & $\leq 10$ & $\leq 20$ \\
    \hline
    SEN1 & \textbf{98.7} & \textbf{99.6} & \textbf{99.6} & \textbf{99.8} \\
    Radarsat & 88.4 & 95.8 & 95.8 & 95.8 \\
   RCM & 59.6 & 74.4 & 75.6 & 77.1 \\
    GF3 & 38.0 & 44.5 & 44.7 & 45.8 \\
    ALOS & 82.0 & 85.3 & 85.3 & 85.6 \\
    HOPC* & 98.4 & 98.7 & 98.7 & 98.9 \\
    \hline
  \end{tabular}}
  \end{subtable}
\end{table}

\subsection{The influence of training data}
In this section, we compared the matching performance of MFN trained on different datasets on the same test set, using HOPC as a baseline for comparison. The experimental results are shown in \cref{tab:result1}. We found that inconsistent sources of training and testing data lead to a decrease in the model’s matching performance. As illustrated in \cref{tab:result1} (a) and (b), Model trained on TerraSAR data exhibits the best matching performance on the TerraSAR test set, while other methods also show some level of matching capability but not as effective as models trained on data from the same source. Additionally, from \cref{tab:result1}(c) and (d), it can be observed that models trained on different source data sometimes perform even worse than HOPC. This series of experiments highlights the importance of training deep learning methods on data from different satellites in the research of optical-SAR image matching. Furthermore, existing deep learning methods need to address domain adaptation issues.
\section{Conclusion}
In this paper, we proposed the 3MOS dataset for Optical-SAR image matching, which contains SAR images from diverse satellites, featuring varying spatial resolutions and scenes. The 3MOS dataset poses challenges as existing methods struggle to achieve high matching accuracy across all data instances. Additionally, we utilize a Multi-Scale Feature Network (MFN) as a representative deep learning model, highlighting the benefits of deep learning approaches on the 3MOS dataset. However, we also recognize the imperative of addressing domain adaptation issues within current neural network architectures. The limitations lie in the uneven distribution of the dataset, the relatively crude division method of data scenes, and there is still considerable room for improvement.

\bibliographystyle{unsrt}  
\bibliography{references}
\newpage
\appendices

\section{The details of original image pairs}
\label{appendix:appendix1}
We used a total of 14 pairs of image data from 6 SAR satellites and Google Earth to construct the 3MOS dataset. The details of these original image data are shown in \cref{tab:original}, including the SAR data sources, image locations, geographic coordinates at the center of the images, polarization modes, imaging time and image size.
\begin{table}[h]
  \caption{The details of original image pairs.
  }
  \label{tab:original}
  \centering
  \scalebox{0.7}{
  \begin{tabular}{ccccccc} 
  \toprule
\textbf{Index} & \textbf{SAR sources} & \textbf{Locations} & \textbf{Center coordinates} & \textbf{Polarization modes} & \textbf{Imaging time} & \textbf{Image size} \\
\midrule
1 & TerraSAR & hainan in China & 109.5E 18.4N & HH & 2014/11/23 & $66800\times 73200$ \\
2 & TerraSAR & Vietnam & 109.0E 12.0N & HH & 2014/12/10 & $47600\times 66400$\\
3 & TerraSAR & Vietnam & 108.8E 11.8N & HH & 2014/11/29 & $68800 \times 41200$ \\
4 & GF3 & hunan in China & 111.8E 29.0N & HH & 2016/10/26 & $18450 \times 19490$ \\
5 & Radarsat & shanxi in China & 109.0E 34.6N & HH & 2012/01/03 & $14423 \times 10580$ \\
6 & ALOS & gansu in China & 93.5E 40.4N & HV & 2015/08/29 & $6324 \times  10010$ \\
7 & RCM & sicuan in China & 103.6E 30.6N & HV & 2023/08/09 & $12555 \times 13657$ \\
8 & SEN1 & Zimbabwe & 32.9E 23.7S & VH & 2023/02/01 & $29192 \times 23035$ \\
9 & SEN1 & Antarctica & 59.7W 64.9S & HH & 2023/06/23 & $32120 \times 29101$ \\
10 & SEN1 & Iraq & 45.5E 31.5N & VH & 2023/02/01 & $28718 \times 21426$ \\
11 & SEN1 & the USA & 88.5W 41.9N & HV & 2024/02/04 & $11237 \times 17806$ \\
12 & SEN1 & Netherlands & 34.9E 52.4N & VV & 2023/06/14 & $29659 \times 22211$ \\
13 & SEN1 & jiangxi in China & 114.2E 27.8N & VH & 2017/06/28 & $28846 \times 21944 $\\
14 & SEN1 & hunan in China & 112.2E 27.9N & VH & 2017/07/05 & $28581 \times 21590$\\

  \bottomrule
  \end{tabular}}
\end{table}
\section{Image Scene Classification}
\label{appendix:appendix2}
To facilitate the evaluation of the accuracy and robustness of image matching methods in different scenarios, the 3MOS dataset has been categorized based on the characteristics of different scenes in optical and SAR images. In this section, we first introduce more information about the image scene classification process. Then, we provide detailed descriptions of the characteristics of different scenes and showcase more example image pairs.
\begin{figure}[h]
  \centering
  \includegraphics[height=7cm]{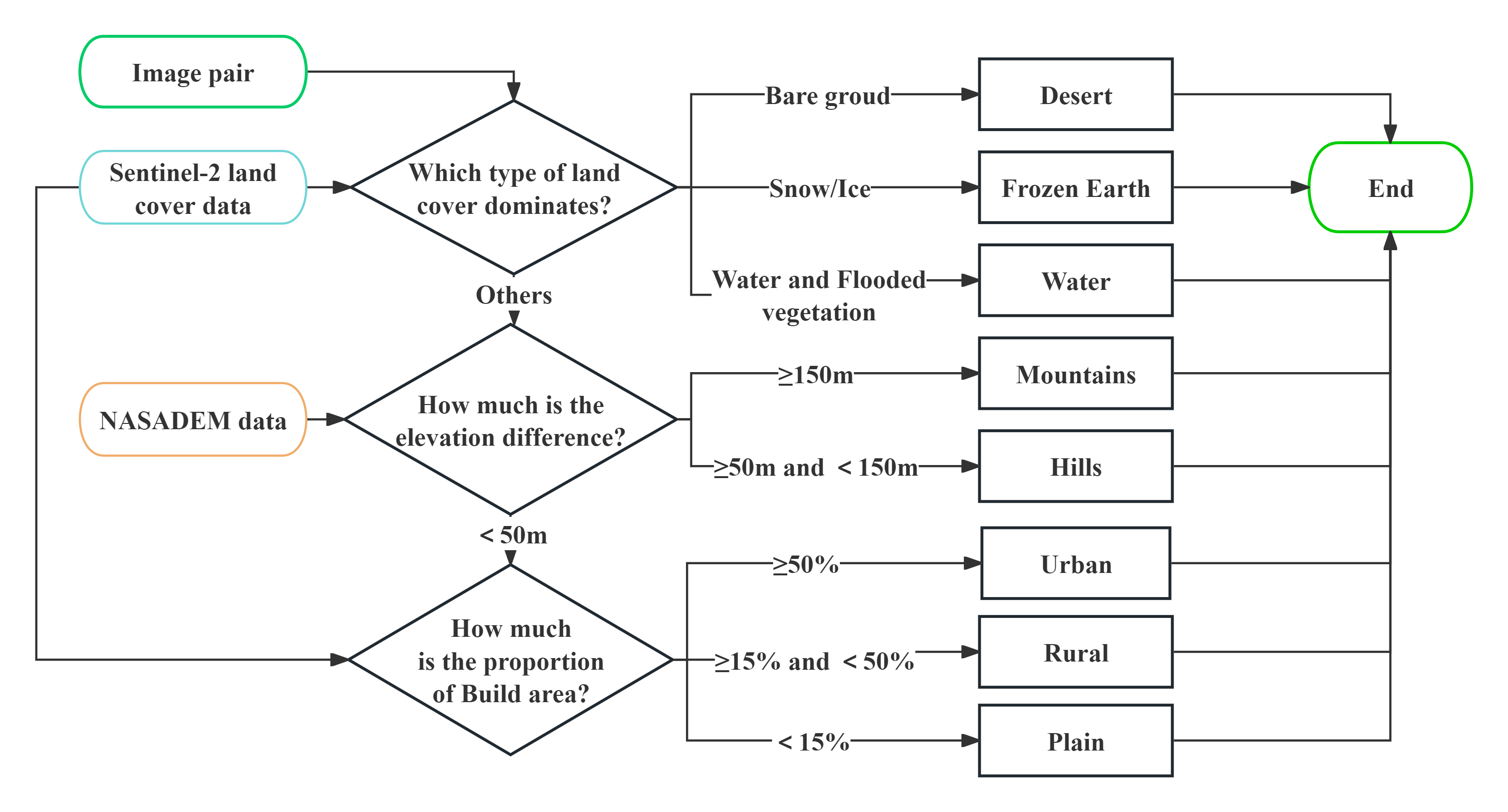}
  \caption{The detailed flowchart of image scene classification
  }
  \label{fig: classification1}
\end{figure}
\subsection{The detail of image scene classification}

During the classification process, we utilize the NASADEM data to differentiate various terrains and the Sentinel-2 Land cover data to identify the primary land cover categories within the images. \textbf{NASADEM} is an updated version of the Digital Elevation Model (DEM) and related products derived from data collected during the Shuttle Radar Topography Mission (SRTM). It utilizes interferometric SAR data from SRTM, which has been reprocessed using an optimized hybrid processing technique to produce the data products. The resolution of this dataset is 30 meters. The \textbf{Sentinel-2 Land cover} data provides a worldwide representation of land use and land cover (LULC) through imagery from ESA Sentinel-2 at a 10m resolution. Generated annually, it utilizes Impact Observatory’s deep learning AI model for land classification, trained on billions of human-labeled image pixels sourced from the National Geographic Society. Predictions encompass nine LULC classes: Water, Trees, Flooded vegetation, Crops, Built Area, Bare ground, Snow/Ice, Clouds, and Rangeland.

In \cref{fig: classification1}, we present  a more detailed process for classifying image scenes. Initially, we identify images belonging to three categories: Desert, Frozen earth, and Water, based on the dominant land cover type. Next, we determine Mountains and Hills scenes according to the elevation difference.  Lastly, we further classify the remaining images into Urban areas, Rural areas, and Plains scenes based on the proportion of built areas.

\subsection{The characteristics of different scenes}

Images of different scenes exhibit distinct characteristics, and variations in terrain and material radiative properties contribute to significant differences between optical and SAR imagery. In this section, we will offer comprehensive descriptions of various scenes and present additional examples to enhance readers’ intuitive and profound comprehension of the 3MOS dataset.

\begin{figure}[h]
  \centering
  \includegraphics[height=3cm]{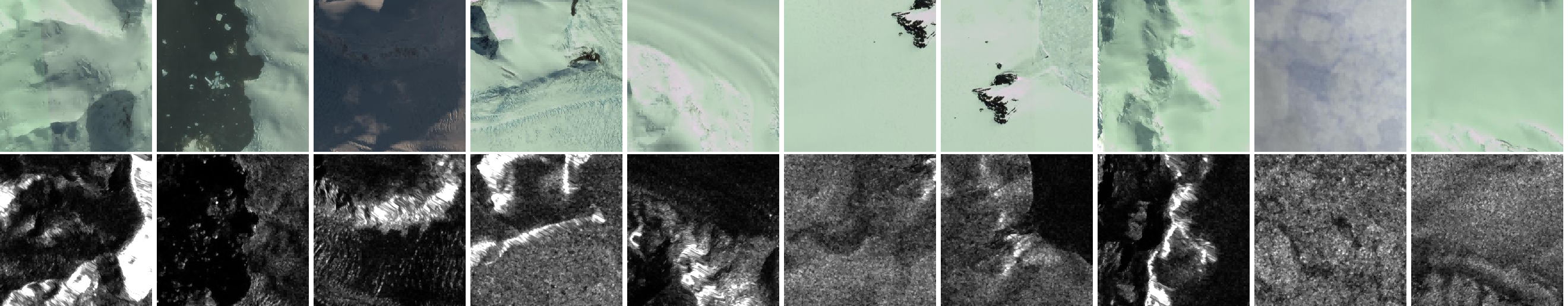}
  \caption{Frozen earth(SEN1) 
  }
  \label{fig:SEN1_frozen}
\end{figure}

\textbf{Frozen earth}(\cref{fig:SEN1_frozen}):  In this category, optical images are often covered by snow and ice with fewer texture features, while SAR images can reveal terrain features beneath the snow and ice. The characteristics reflected in the two types of images are completely different, leading to greater difficulty in matching.

\begin{figure}[h]
  \centering
  \begin{subfigure}[h]{1\textwidth} 
    \centering
    \includegraphics[height=3cm]{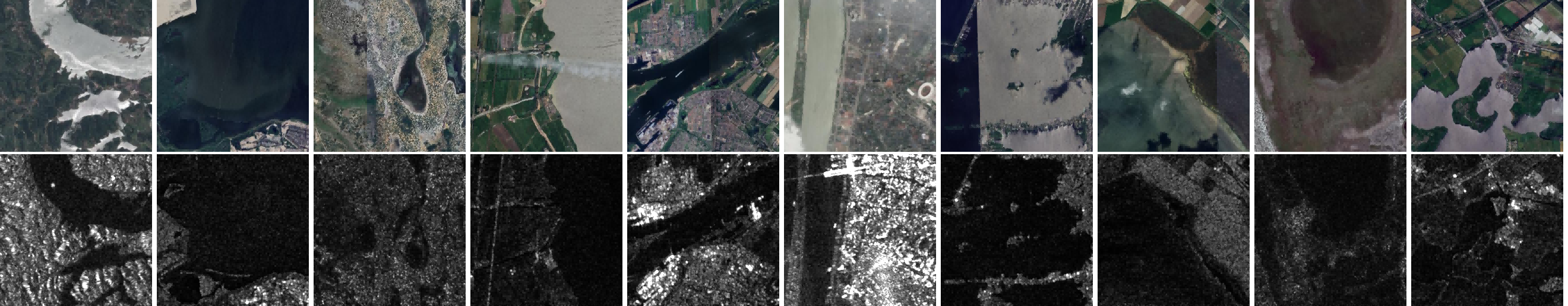}
    \caption{SEN1 Water}
    \label{fig:su1}
  \end{subfigure}
  
  \begin{subfigure}[h]{1\textwidth} 
    \centering
    \includegraphics[height=3cm]{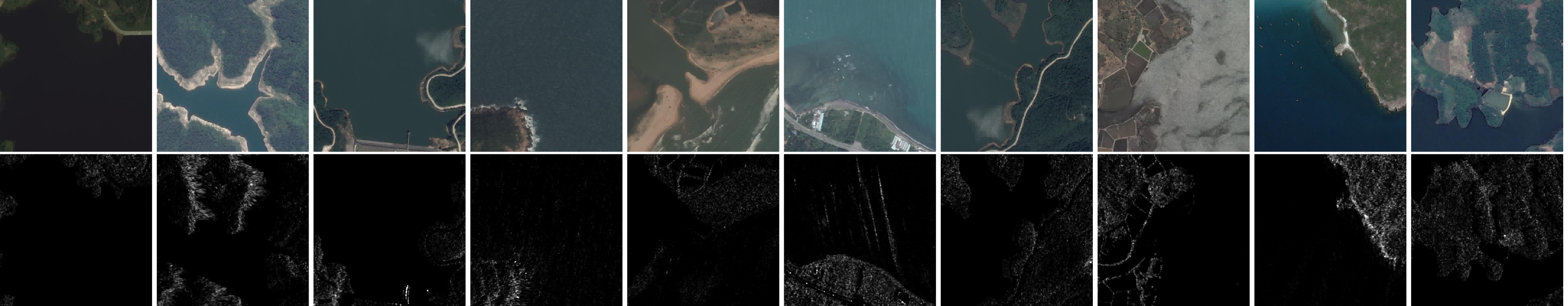}
    \caption{TerraSAR Water}
    \label{fig:sub21}
  \end{subfigure}

  \begin{subfigure}[h]{1\textwidth} 
    \centering
    \includegraphics[height=3cm]{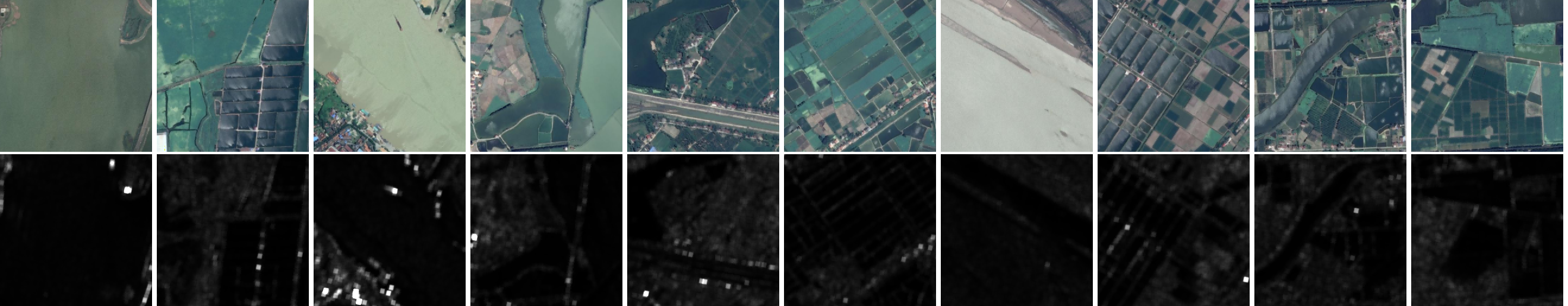}
    \caption{GF3 Water}
    \label{fig:sub12}
  \end{subfigure}
  
  \caption{Water}
  \label{fig:water}
\end{figure}

\textbf{Water}(\cref{fig:water}): In SAR images, most of the electromagnetic waves incident on the water surface are not received by SAR sensors due to specular reflection. Consequently, the water areas of SAR images contains relatively limited information, except for significant targets such as ships. In contrast, the water area in optical images reflects richer information, varying with factors such as illumination, season, and aquatic vegetation. Additionally, water areas are mostly characterized by repetitive texture or no texture, matching images in this area can only utilize the small portion of land area information.

\begin{figure}[h]
  \centering
  \begin{subfigure}[h]{1\textwidth} 
    \centering
    \includegraphics[height=3cm]{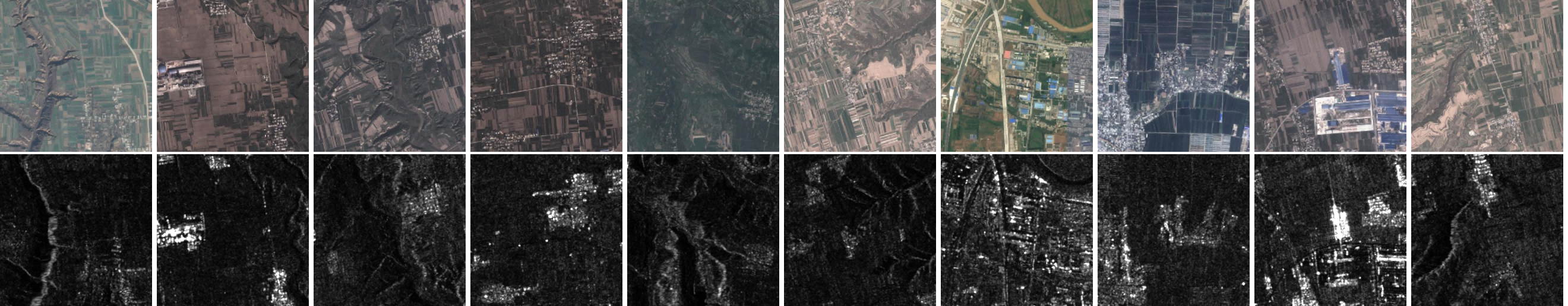}
    \caption{Radarsat Hills}
  \end{subfigure}

 \begin{subfigure}[h]{1\textwidth} 
    \centering
    \includegraphics[height=3cm]{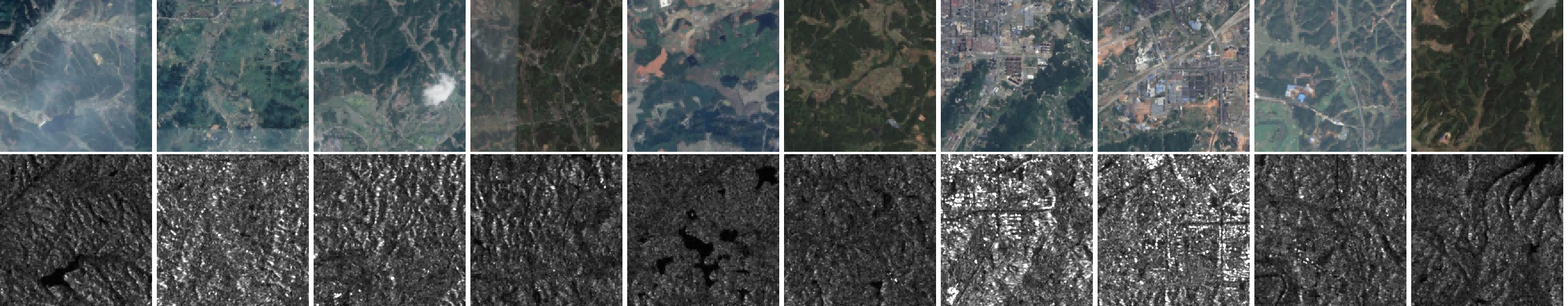}
    \caption{SEN1 Hills}
  \end{subfigure}

    \begin{subfigure}[h]{1\textwidth} 
    \centering
    \includegraphics[height=3cm]{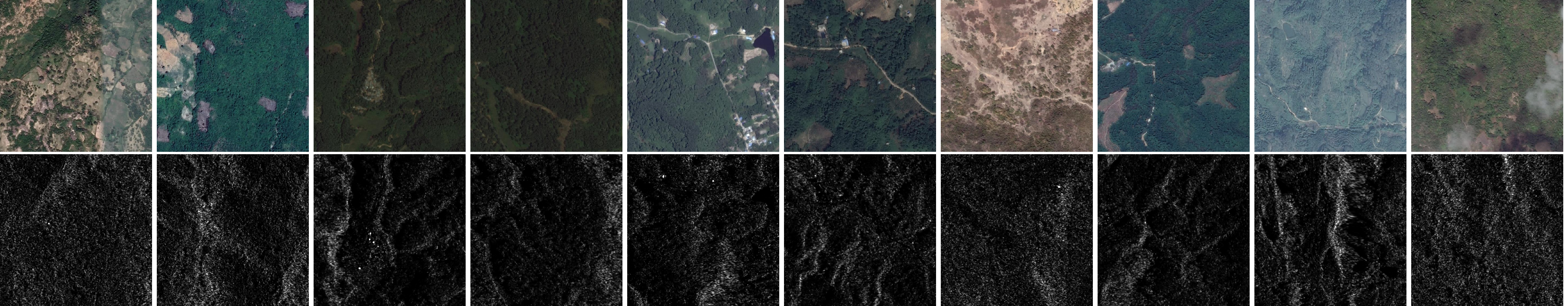}
    \caption{TerraSAR Hills}
  \end{subfigure}

  \caption{Hills}
  \label{fig:hills}
\end{figure}

\begin{figure}[h]
  \centering
  \begin{subfigure}[h]{1\textwidth} 
    \centering
    \includegraphics[height=3cm]{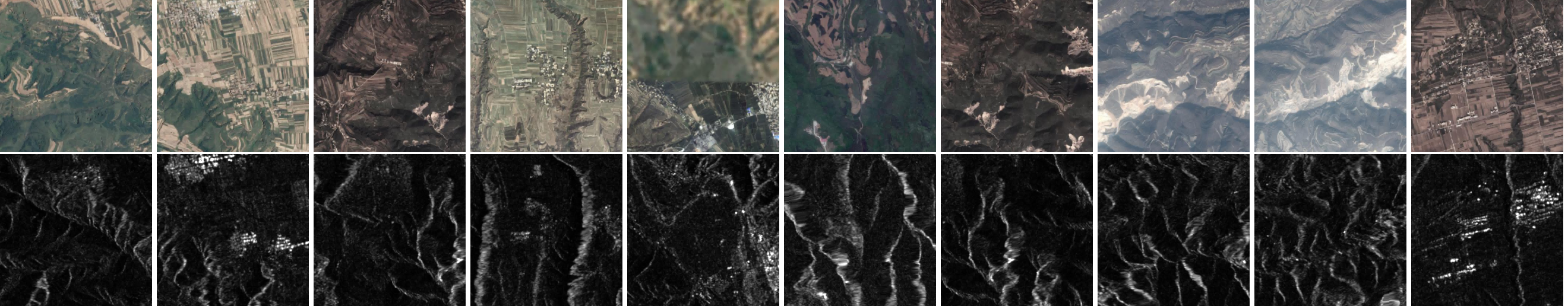}
    \caption{Radarsat Mountains}
  \end{subfigure}

 \begin{subfigure}[h]{1\textwidth} 
    \centering
    \includegraphics[height=3cm]{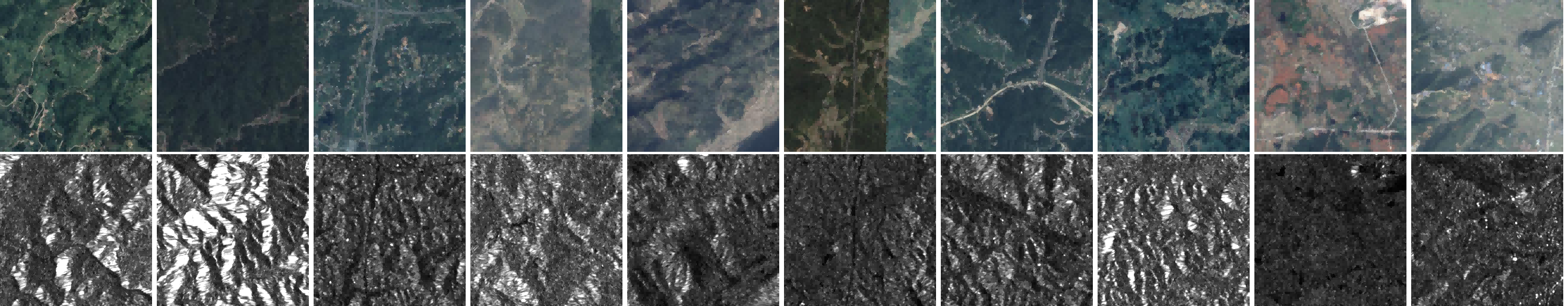}
    \caption{SEN1 Mountains}
  \end{subfigure}

    \begin{subfigure}[h]{1\textwidth} 
    \centering
    \includegraphics[height=3cm]{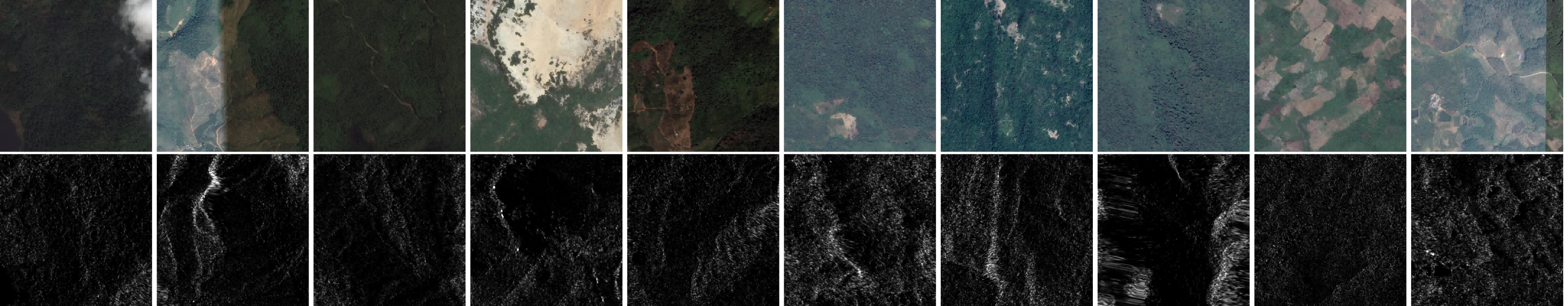}
    \caption{TerraSAR Mountains}
  \end{subfigure}

  \caption{Mountains}
  \label{fig:Mountains}
\end{figure}

\textbf{Hills}(\cref{fig:hills}) and \textbf{Mountains}(\cref{fig:Mountains}): In these areas, SAR imaging often witnesses geometric distortions such as perspective shortening and layover, and due to the greater elevation differences in mountainous areas, these distortions are more pronounced compared to hilly areas. Additionally, optical images mainly observe surface plants, while SAR images can reveal terrain features beneath the plants.

\begin{figure}[h]
  \centering
  \begin{subfigure}[h]{1\textwidth} 
    \centering
    \includegraphics[height=3cm]{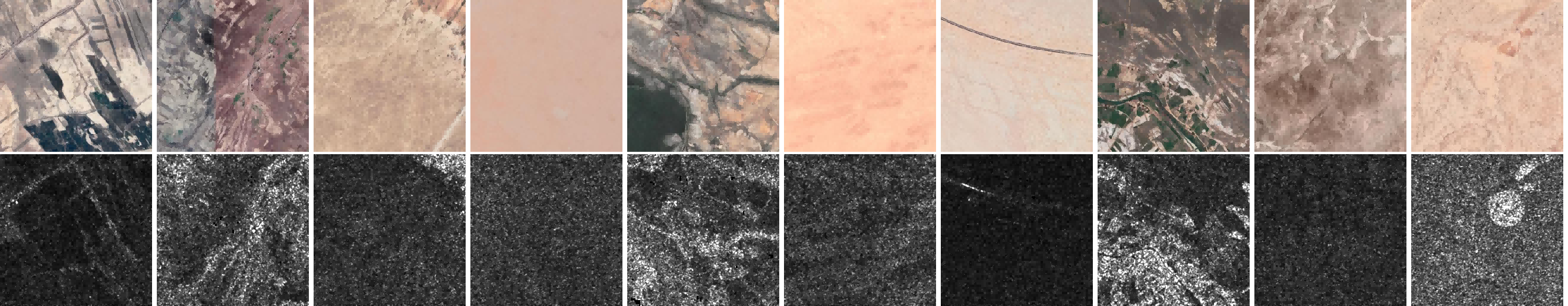}
    \caption{SEN1 Desert}
  \end{subfigure}
  
  \begin{subfigure}[h]{1\textwidth} 
    \centering
    \includegraphics[height=3cm]{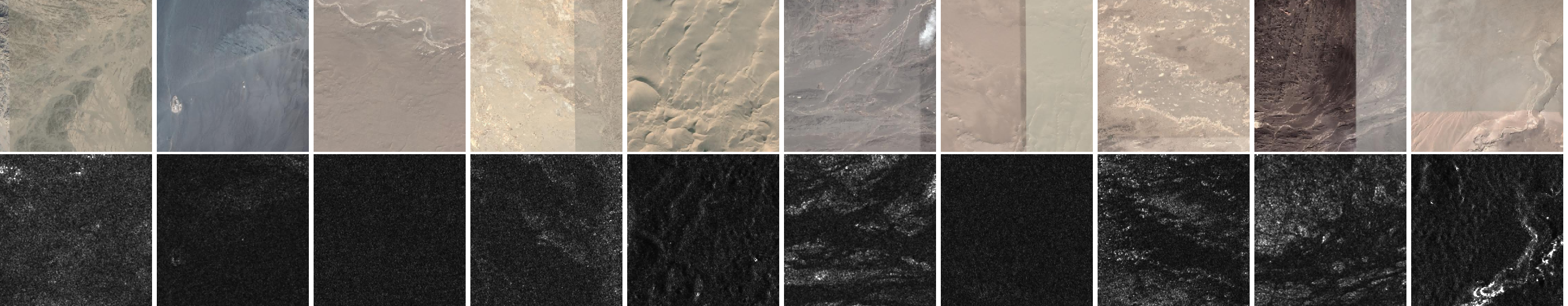}
    \caption{ALOS Desert}
    \label{fig:sub211}
  \end{subfigure}

  \caption{Desert}
  \label{fig:desert}
\end{figure}

\textbf{Desert}(\cref{fig:desert}): SAR images of desert areas are usually characterized by low backscatter because of the sparse vegetation and lack of moisture in the terrain. Unique geological features such as sand patterns and rock formations can be highlighted in SAR images of deserts due to their surface roughness. Due to the influence of wind, images in this area often undergo significant changes over time.

\begin{figure}[h]
  \centering

   \begin{subfigure}[h]{1\textwidth} 
    \centering
    \includegraphics[height=3cm]{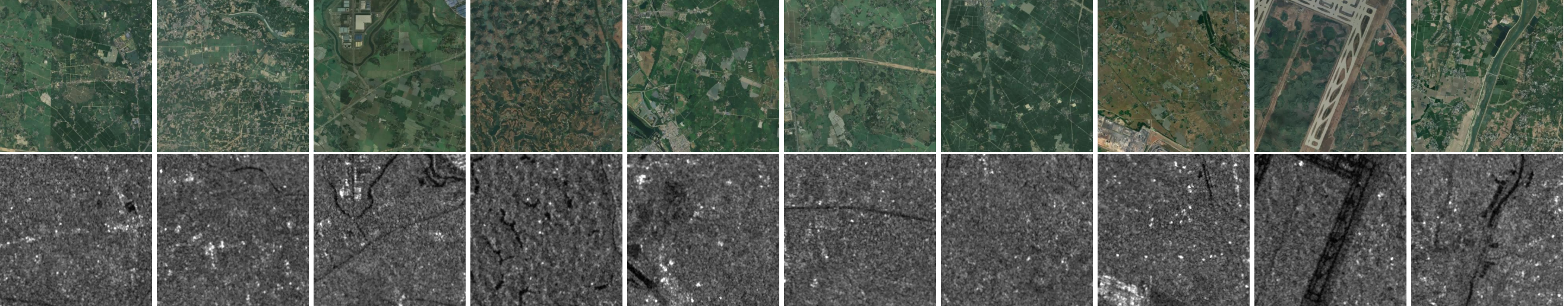}
    \caption{RCM Plains}
  \end{subfigure}
  
 \begin{subfigure}[h]{1\textwidth} 
    \centering
    \includegraphics[height=3cm]{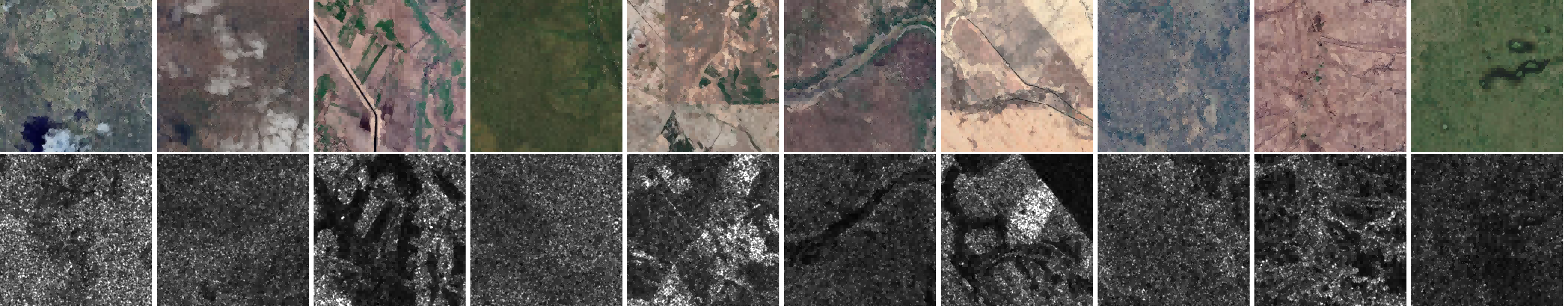}
    \caption{SEN1 Plains}
  \end{subfigure}

   \begin{subfigure}[h]{1\textwidth} 
    \centering
    \includegraphics[height=3cm]{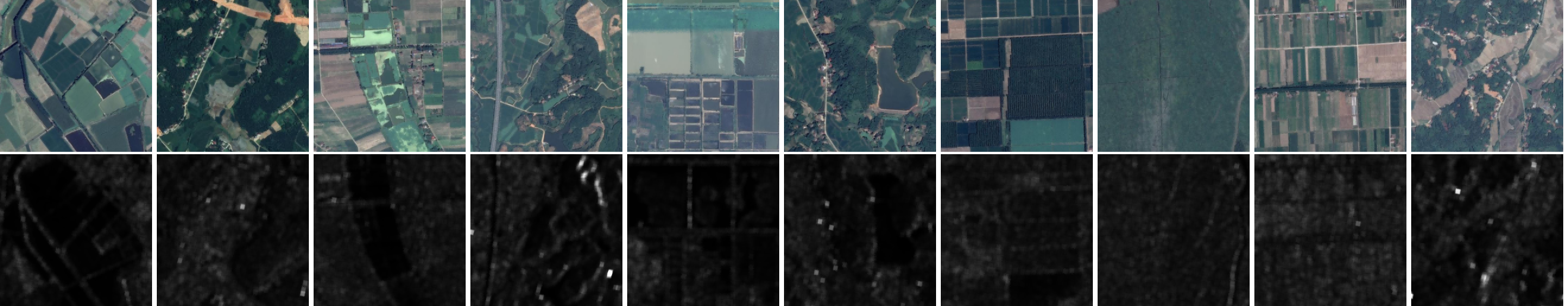}
    \caption{GF3 Plains}

  \end{subfigure}
  
    \begin{subfigure}[h]{1\textwidth} 
    \centering
    \includegraphics[height=3cm]{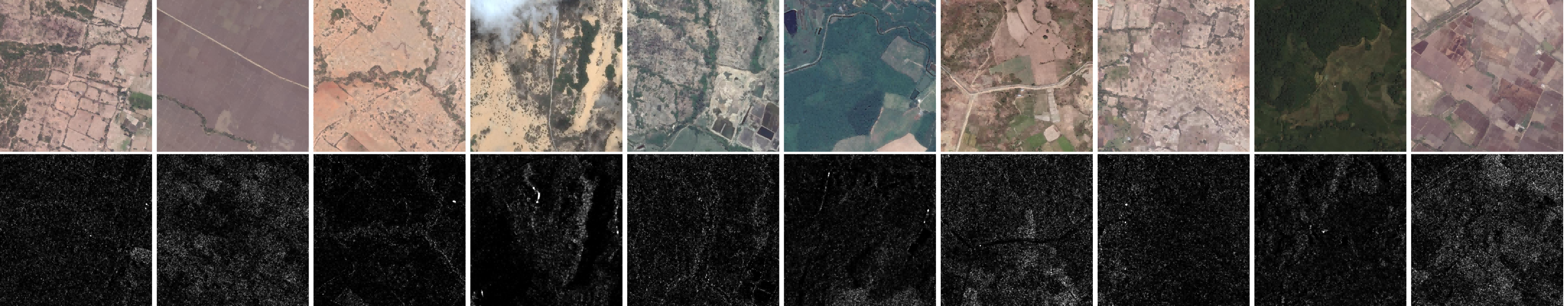}
    \caption{TerraSAR Plains}
  \end{subfigure}

  \caption{Plains}
  \label{fig:Plain}
\end{figure}

\textbf{Plains}(\cref{fig:Plain}) In optical images, plains appear as flat, expansive areas with varying colors depending on vegetation cover, soil type, and they typically exhibit a smooth texture. The uniformity often seen in plains with the naked eye can appear varied and textured in SAR images due to differences in soil composition, vegetation density, and moisture levels.

\begin{figure}[h]
  \centering
  \begin{subfigure}[h]{1\textwidth} 
    \centering
    \includegraphics[height=3cm]{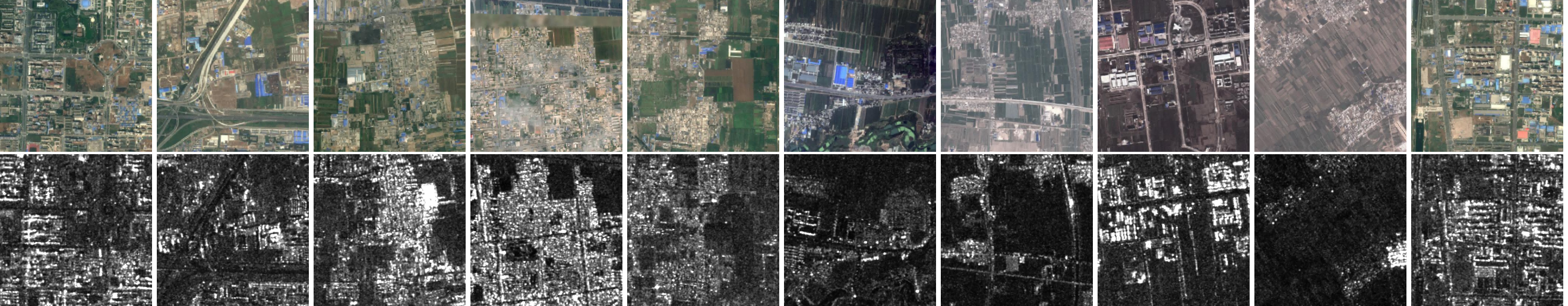}
    \caption{Radarsat Urban areas}
    \label{fig:sub1}
  \end{subfigure}
  
  \begin{subfigure}[h]{1\textwidth} 
    \centering
    \includegraphics[height=3cm]{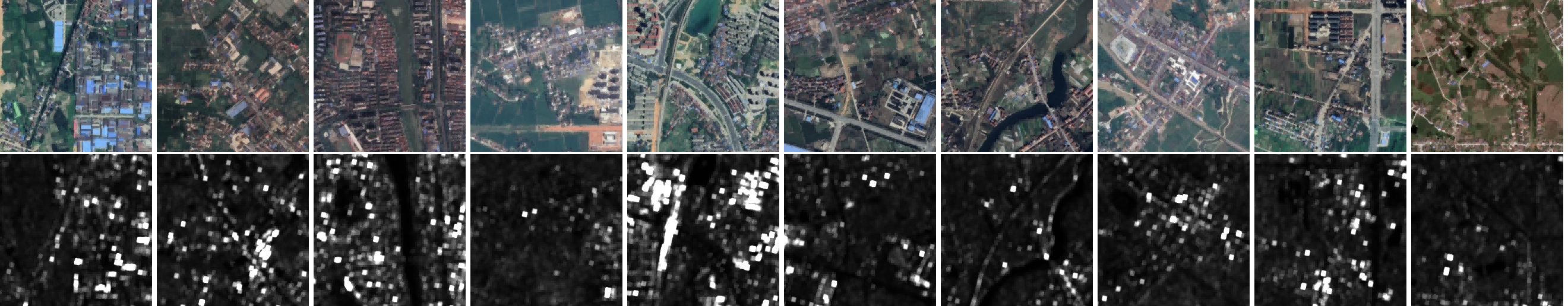}
    \caption{GF3 Urban areas}
    \label{fig:sub212}
  \end{subfigure}

 \begin{subfigure}[h]{1\textwidth} 
    \centering
    \includegraphics[height=3cm]{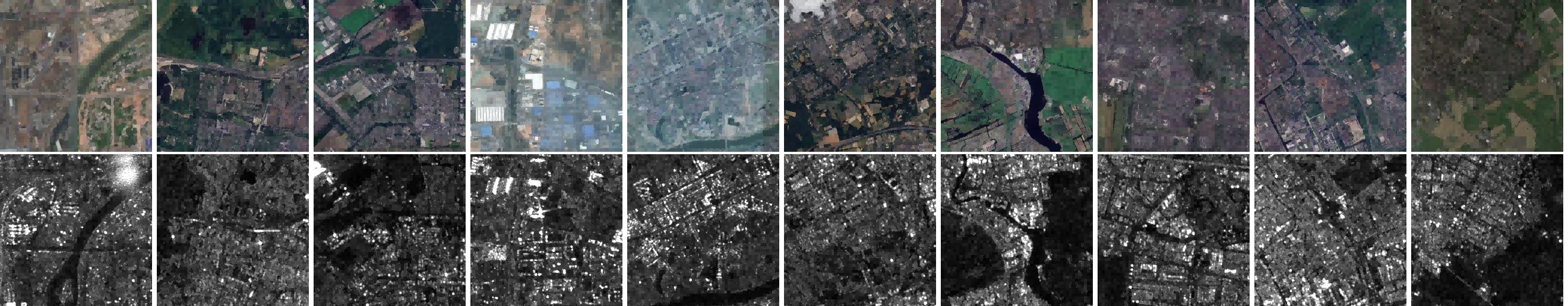}
    \caption{SEN1 Urban areas}
  \end{subfigure}

      \begin{subfigure}[h]{1\textwidth} 
    \centering
    \includegraphics[height=3cm]{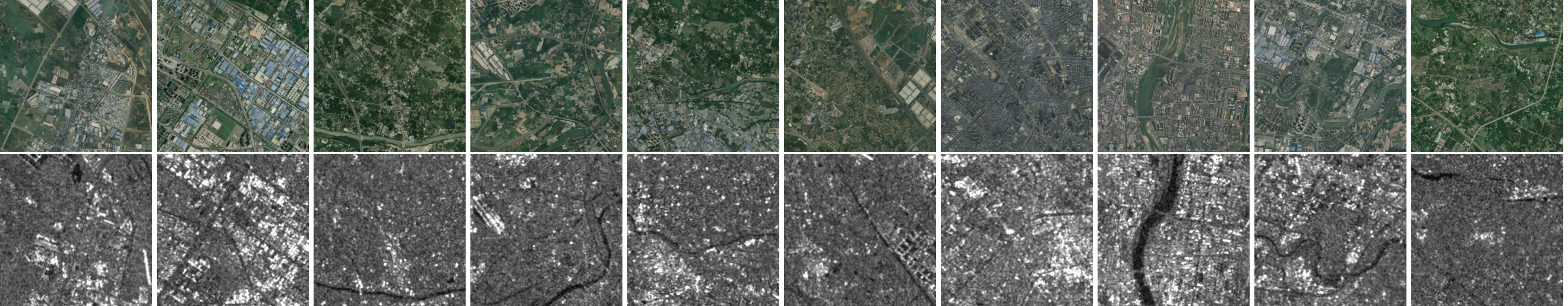}
    \caption{RCM Urban areas}
    \label{fig:sub5}
  \end{subfigure}
  
    \begin{subfigure}[h]{1\textwidth} 
    \centering
    \includegraphics[height=3cm]{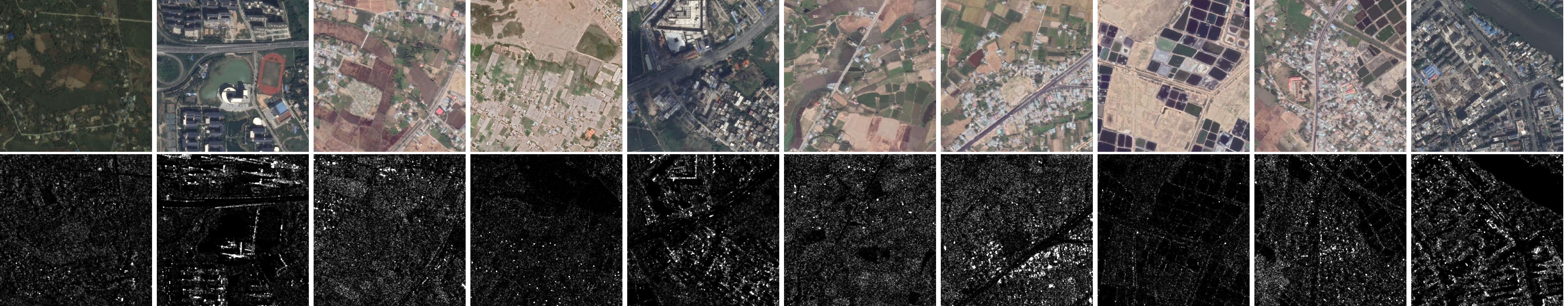}
    \caption{TerraSAR Urban areas}
  \end{subfigure}

  \caption{Urban areas}
  \label{fig:urban}
\end{figure}

\textbf{Urban}(\cref{fig:urban}): Urban areas in SAR images typically exhibit strong scattering, resulting in numerous strong scattering points and higher levels of noise. Additionally, built-up areas can experience significant geometric distortions in range imaging, with tall buildings even appearing inverted from top to bottom. The imaging geometry and grayscale characteristics of urban areas in SAR images differ significantly from those in optical images.

\begin{figure}[h]
  \centering
  \begin{subfigure}[h]{1\textwidth} 
    \centering
    \includegraphics[height=3cm]{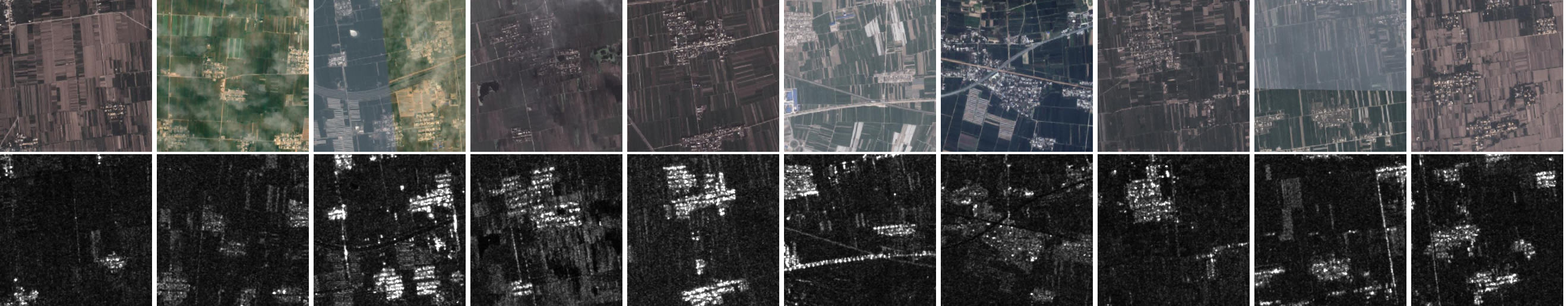}
    \caption{Radarsat Rural areas}
  \end{subfigure}
  
  \begin{subfigure}[h]{1\textwidth} 
    \centering
    \includegraphics[height=3cm]{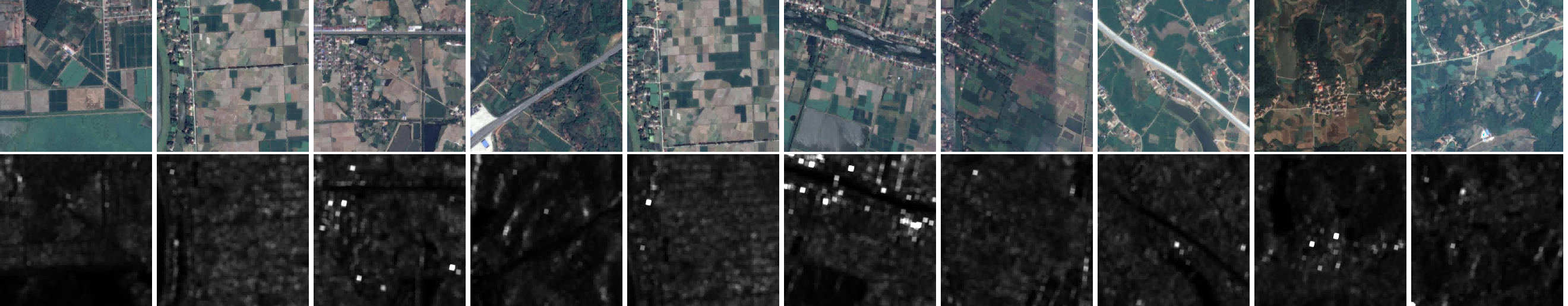}
    \caption{GF3 Rural areas}
    \label{fig:sub112}
  \end{subfigure}

 \begin{subfigure}[h]{1\textwidth} 
    \centering
    \includegraphics[height=3cm]{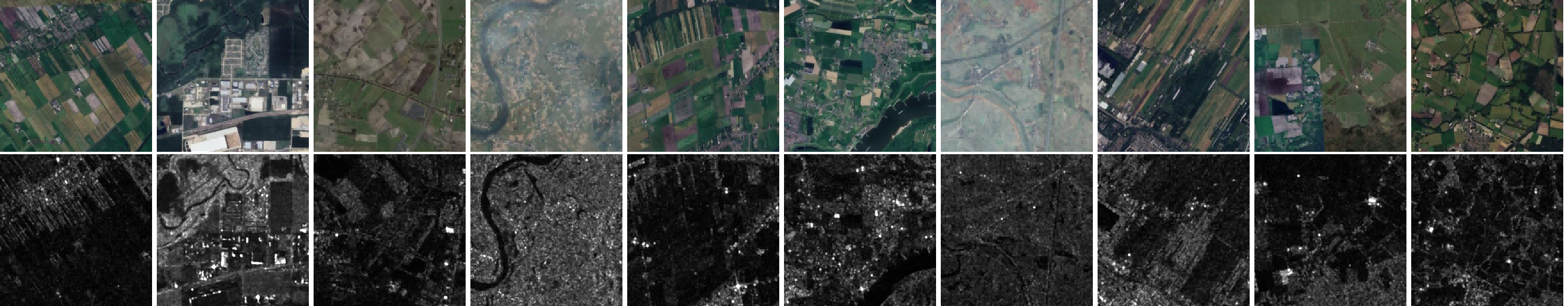}
    \caption{SEN1 Rural areas}

  \end{subfigure}

      \begin{subfigure}[h]{1\textwidth} 
    \centering
    \includegraphics[height=3cm]{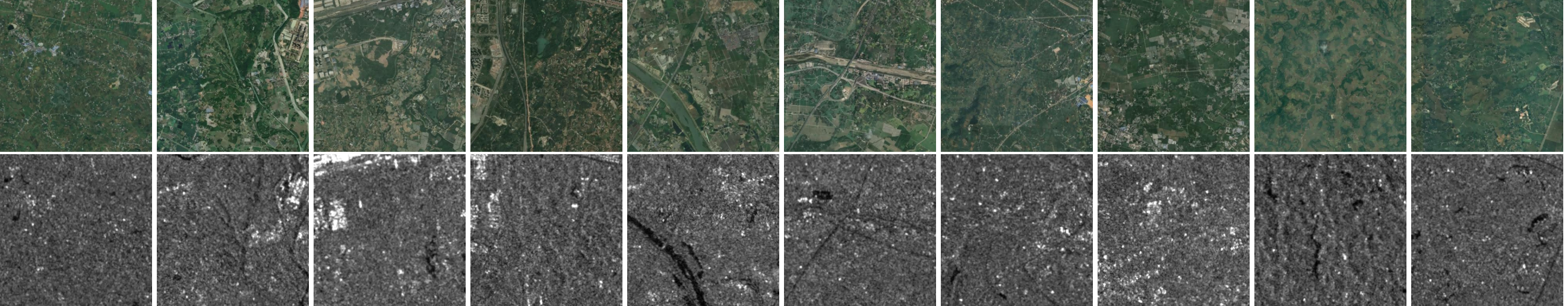}
    \caption{RCM Rural areas}

  \end{subfigure}
  
    \begin{subfigure}[h]{1\textwidth} 
    \centering
    \includegraphics[height=3cm]{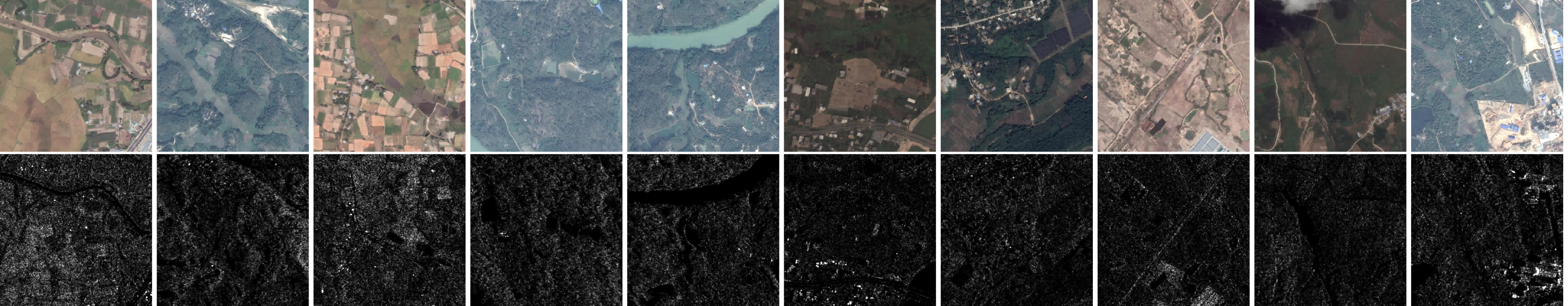}
    \caption{TerraSAR Rural areas}
  \end{subfigure}

  \caption{Rural areas}
  \label{fig:Rural}
 \end{figure}

\textbf{Rural}(\cref{fig:Rural}): Rural areas, situated between plains and urban centers, typically contain a small proportion of human-made structures. In such areas, SAR images may exhibit some degree of strong scattering and geometric distortion, while imaging characteristics in other regions are similar to those of plains.


\end{document}